\pdfoutput=1  %
\documentclass[11pt]{article}
\usepackage[margin=1in]{geometry}
\usepackage{natbib}                 %
\usepackage[utf8]{inputenc}
\usepackage[T1]{fontenc}
\usepackage[hidelinks]{hyperref}
\usepackage{url}
\usepackage{adjustbox}
\usepackage{booktabs}
\usepackage{float}
\newif\ifartifactlink\artifactlinkfalse
\newcommand{\artifacturl}{https://github.com/avalliappan-nvidia/windowed-mtp-b200/tree/master}
\makeatletter
\newenvironment{widefigure}
  {\if@twocolumn\begin{figure*}[t]\else\begin{figure}[!htbp]\fi}
  {\if@twocolumn\end{figure*}\else\end{figure}\fi}
\newenvironment{widetable}
  {\if@twocolumn\begin{table*}[t]\else\begin{table}[!htbp]\fi}
  {\if@twocolumn\end{table*}\else\end{table}\fi}
\makeatother
\usepackage{amsfonts}
\usepackage{amsmath}
\usepackage{amssymb}
\usepackage{nicefrac}
\usepackage{microtype}
\usepackage{graphicx}
\usepackage{multirow}
\usepackage{xcolor}
\usepackage{subcaption}
\usepackage{tikz}
\usetikzlibrary{positioning,decorations.pathreplacing,calc}

\graphicspath{{figs/}}

\newcommand{\tdraft}{t_{\mathrm{draft}}}
\newcommand{\tverify}{t_{\mathrm{verify}}}
\newcommand{\tfwd}{t_{\mathrm{draft}}^{\mathrm{fwd}}}
\newcommand{\tctx}{t_{\mathrm{draft}}^{\mathrm{ctx}}}
\newcommand{\AL}{\mathrm{AL}}
\newcommand{\wperturb}{\mathrm{win\_perturb}}
\artifactlinktrue  %

\title{Windowed-MTP: Removing the Full-Context Draft-KV Tax at Million-Token Context}
\author{Alagappan Valliappan\\ NVIDIA\\ \texttt{avalliappan@nvidia.com}}
\date{}

\begin{document}

\maketitle

\begin{abstract}
Speculative decoding accelerates autoregressive generation by having a
\emph{draft} propose tokens that a \emph{target} verifies in parallel. Frontier models
increasingly ship a built-in Multi-Token-Prediction (MTP) draft head, assumed to be
cheap beside verification. \textbf{A shipped full-attention MTP head can become
disproportionately expensive beside a hybrid target at long context.} The head attends
over the full KV cache for each of $\gamma$ draft forwards per decode step, while only
$9$--$25\%$ of target layers retain full attention, so a single draft layer can
issue context reads at a rate comparable to the whole target. On hard, low-acceptance
workloads, deep native MTP becomes \emph{net-negative}---slower than no speculation.

We apply a StreamingLLM-style window plus attention sink to the \emph{draft} only
(\textbf{Windowed-MTP}); target verification remains full-context, so the target
distribution is preserved. The intervention is training-free and drop-in.
On three models from two hybrid families (Gated DeltaNet and Mamba2) at 1M context on one
GPU in SGLang~\citep{zheng2024sglang}, it provides an acceptance-free
\textbf{1.28--1.44$\times$ per-decode-step speedup} over native MTP.
The speedup grows with context; at matched acceptance,
TPOT improves by the same factor. The window removes most of the draft's working set, and
a compact ring buffer converts that reduction into physical capacity: it reclaims the
otherwise dead draft pool---a measured 7.7--11.1\% of total KV at 1M---making room for one
additional concurrent 1M request in the Qwen-35B batch sweep.
\end{abstract}

\section{Introduction}
\label{sec:intro}

Speculative decoding (SD)~\citep{leviathan2023speculative,chen2023accelerating}
reduces autoregressive latency by proposing $\gamma$ candidate tokens and verifying them
in one batched target forward. Recent frontier models bake the draft into the target as a
lightweight Multi-Token-Prediction (MTP) head~\citep{deepseekv3,qwen35,qwen36}, avoiding
a separate draft model.

Standard SD analyses model a draft forward as costing a fixed, small fraction of a
target forward~\citep{leviathan2023speculative}. For a one-layer MTP head that is a fair
assumption at short context, but at long context the fraction is neither small nor
fixed.
Hybrid and linear-attention targets (GDN, Mamba2) reduce verification's KV cost, whereas
the MTP heads on the models we study retain full softmax/GQA attention. At
$\gamma{=}6$, the single draft layer reads the full KV cache six times per step. The
targets we test have only $8$--$12$ full-attention layers, each read once, so the draft
does $50$--$75\%$ as much KV-read work as the entire target. This
\emph{long-context draft tax} can erase, and on hard workloads invert, SD's speedup. At
the $d{=}7$ operating point, native MTP falls \emph{below} the dense baseline on hard 1M tasks
($0.86\times$ on Qwen-122B code QA, $0.58\times$ on Nemotron CWE;
Table~\ref{tab:hero}).

A draft only needs to be a \emph{good enough guesser}; it need not reread the whole
million-token history. We therefore apply a StreamingLLM-style sliding window plus
attention sink~\citep{xiao2024streamingllm} to the \textbf{draft's attention only},
leaving verification at full attention. The draft working set becomes
$W+\text{sink}$ tokens (4K is $0.4\%$ of 1M), with no training, extra parameters, or
distillation, and a ring buffer converts the smaller working set into physical capacity.

\paragraph{Contributions.}
\begin{enumerate}
  \item We quantify the \textbf{long-context draft-attention tax} in built-in MTP
  heads. At 1M ($\gamma{=}6$), the draft alone takes $61$--$102\%$ as long as the target's
  verify pass; with fixed speculative bookkeeping, speculation's overhead is
  $92$--$138\%$ of it (Table~\ref{tab:iterphase}). We give a per-decode-step model that predicts when a deep native
  draft reaches parity with, or falls below, dense decoding (\S\ref{sec:motivation}).
  \item We retrofit a \textbf{training-free window+sink onto the draft attention}
  (\textbf{Windowed-MTP}%
  \ifartifactlink\footnote{Reproduction package (SGLang patch, run/config scripts,
  seeded inputs): \url{\artifacturl}. The windowed draft attention is merged
  upstream in SGLang (\href{https://github.com/sgl-project/sglang/pull/32673}{PR~\#32673}).}\fi) of heads that were \emph{trained} at full
  attention, at serving time and with the verifier unchanged (\S\ref{sec:method}). On
  three models from two hybrid families it gives a \textbf{$\mathbf{1.28}$--$\mathbf{1.44\times}$
  acceptance-free per-decode-step speedup at 1M} ($B{=}1$), widening from
  $1.17$--$1.27\times$ at 261K, with magnitude tracking the windowable share of draft cost
  (\S\ref{sec:results}).
  \item We \textbf{measure how windowing changes acceptance, and why}. We sweep the window
  size, break acceptance down by draft position, and measure in live runs how often
  windowing changes the draft's top-1 proposal: it stays the same $86$--$94\%$ of the
  time. Acceptance usually changes little (median windowed-to-native ratio $0.96$ over
  $15$--$16$ prompts per task); the largest task-level median loss is $19\%$. The per-step saving outweighs the loss on every
  model--task cell we measure (\S\ref{sec:mechanism}).
  \item The draft KV pool is a \textbf{measured $7.7$--$11.1\%$ of total KV} at 1M;
  windowing renders all but $W{+}\text{sink}$ of it dead, and a compact
  \textbf{ring buffer converts that into physical capacity} at no quality or speed cost.
  In the Qwen-35B batch sweep, with each arm at its best feasible memory setting, it fits
  a sixth concurrent 1M request where native fits five (\S\ref{sec:results}).
\end{enumerate}

\section{Background and related work}
\label{sec:background}

\paragraph{Speculative decoding and MTP heads.}
SD~\citep{leviathan2023speculative,chen2023accelerating} pairs a draft with a
target verifier. Draft designs range from a separate small
model~\citep{leviathan2023speculative} to self-drafting with extra prediction
heads on the target itself~\citep{cai2024medusa} to feature-level autoregression
(EAGLE)~\citep{li2024eagle,li2024eagle2}. Frontier models such as
DeepSeek-V3~\citep{deepseekv3} and Qwen3.5/3.6~\citep{qwen35,qwen36} ship a built-in MTP head, which we target here. Acceptance length $\AL$ (mean tokens accepted
per verify step) governs the speedup; prior work optimizes $\AL$ via better draft
alignment or tree drafts~\citep{li2024eagle2}. Our work is orthogonal: we reduce
the draft's \emph{cost} at long context without touching the verifier.

\paragraph{Long-context SD.}
MagicDec~\citep{chen2024magicdec} observes that at long context and large batch
the SD bottleneck shifts to KV loading, and uses a sparse KV cache in a self-speculative decoding setup.
We share that diagnosis but focus on the
\emph{built-in MTP draft head} of shipping models, and analyze when windowing
preserves or trades acceptance.
Closest to us, LongSpec~\citep{yang2026longspec} \emph{trains} a dedicated draft
with a constant-size KV cache and hybrid tree attention (dense targets,
${\le}64$K), while SpecExtend~\citep{cha2026specextend} is training-free but relies
on a \emph{separate} draft model with a cross-model retrieval KV policy, at
shorter long-context regimes than the 1M we target.
We instead window the \emph{already-shipped} MTP head of hybrid-attention
models at 1M, where a cheap verifier \emph{exposes} the draft's $O(S)$ read, inside a
production serving framework (continuous batching, paged/radix KV, CUDA-graph capture).

\paragraph{Streaming attention and context extension.}
StreamingLLM~\citep{xiao2024streamingllm} shows that retaining a few initial
``sink'' tokens plus a recent window preserves fluency under bounded attention;
we repurpose this as a \emph{draft-only} approximation. SpecExtend~\citep{cha2026specextend}
argues StreamingLLM-style windowing struggles on needle retrieval, but that is for a
\emph{separate} small draft whose only route to a distant needle is its own attention. An
MTP head instead conditions on the target's last hidden state, which already carries the
target's full-context retrieval, so windowing the draft's own attention removes much less
(\S\ref{sec:mechanism}). As in any SD scheme, a missed far token costs at most a rejection,
never a wrong answer. Concurrent
work~\citep{eldenk2026attentiondrift} independently reports drafter ``attention drift''
toward recent tokens (EAGLE-3 and MTP heads) and windows the draft at ${\le}32$K as an
evaluation tool alongside a training-time post-norm fix; we make windowing a training-free
1M serving technique with draft-KV reclamation. RoPE~\citep{su2024roformer} with
position interpolation~\citep{chen2023extending} and YaRN~\citep{peng2023yarn} enables the
million-token contexts we study; to show the tax and its fix are not RoPE-specific, we also
evaluate a Mamba2-hybrid~\citep{gu2023mamba,dao2024mamba2} whose attention uses no position
embedding (NoPE).

\paragraph{Drafts windowed at training time.}
Three recent releases ship MTP drafters whose attention is partly or wholly local.
MiMo-V2-Flash~\citep{mimov2flash} is the most direct: its MTP block uses sliding-window
attention with \emph{no} global attention at all, at a $128$-token window, and states the
reason outright---to cut KV-cache and attention cost. Gemma 4~\citep{gemma4} uses a
four-layer drafter with three local layers and one global, and
Inkling~\citep{inkling2026} chains single-layer heads with full or sliding-window
attention over a mostly-SWA backbone at 1M. We read these as corroboration rather than
competition: they show publicly that a draft need not read the whole context.

The delta is train time versus serving time. These releases establish that a drafter
\emph{trained} with local attention works. None addresses whether a head \emph{trained}
at full attention tolerates being windowed at inference---the question facing anyone
serving an already-published checkpoint, who cannot retrain it. That retrofit is our
result, and is why the contribution is aimed at the runtime rather than the model card.
Independently of it, none of the three quantifies the tax or gives a cost model predicting
where speculation inverts.

\paragraph{Serving systems.}
We build on paged-KV serving~\citep{kwon2023vllm} and implement Windowed-MTP in
SGLang~\citep{zheng2024sglang}. The draft's paged block table is reduced to sink
plus recent blocks, a graph-safe change requiring no new attention kernel on either
the FlashInfer or Triton draft backend. Public serving support now provides this
logical sink-plus-recent KV selection for built-in MTP/EAGLE draft decoding. Our
evaluated system additionally bounds the \emph{physical} draft-KV allocation: entries
excluded from the read set are not merely skipped at attention time, but need not remain
resident. We compact them into an $O(n_{\text{sink}}{+}W{+}d)$ ring buffer, freeing HBM
for additional concurrency (\S\ref{sec:results}). The compact-ring remap currently uses
the Triton draft backend.

\section{The long-context draft-attention tax}
\label{sec:motivation}

\paragraph{Cost of one decode step.}
A speculative decode step runs $\gamma$ draft forward passes, proposing $\gamma$ tokens;
the target verifies them in one forward and appends one bonus token, so a step commits
between $1$ and $\gamma{+}1$ tokens, on average the acceptance length $\AL$. We index runs
by this cap, $d\equiv\gamma{+}1$ (SGLang's \texttt{num\_draft\_tokens}), and report
$d\in\{3,5,7\}$, i.e.\ $\gamma\in\{2,4,6\}$ draft forwards. One step takes
\begin{equation}
  \text{step} = \underbrace{\tverify + \tctx}_{c} \;+\; \gamma\,\tfwd,
  \qquad
  \mathrm{TPOT} = \frac{\text{step}}{\AL}.
  \label{eq:tpot}
\end{equation}
$\tverify$ is the target's verify forward over the $d$-token block (the bare verify) plus fixed speculative bookkeeping
(tree build, sampling). The draft phase $\tdraft=\tctx+\gamma\,\tfwd$ has two terms:
$\tctx$, a once-per-step build of the draft's KV index over the whole context, and
$\tfwd$, one draft forward, which rereads the draft's KV. Both grow as $\mathcal{O}(S)$,
so the native draft pays for the full context in both terms. To first order, none of the terms depends on how many tokens a step accepts; their fitted values drift only modestly across workloads
(App.~\ref{app:predict}). Windowing targets the two draft terms.%
\footnote{The target verify also grows with $S$, but windowing leaves the target path
untouched, so it is not a lever here.}

\paragraph{Break-even acceptance length.}
Dense decoding commits one token per step, so its TPOT is its step time. Speculation at
depth $d$ therefore beats dense exactly when its $\AL$ exceeds the \emph{break-even}
$\AL$, the number of dense steps that fit in one native speculative step:
\begin{equation}
  \AL(d) \;>\; \AL_{\text{break-even}}(d) = \frac{\text{step}^{\text{native}}(d)}{\text{step}^{\text{dense}}}.
  \label{eq:breakeven}
\end{equation}
The break-even $\AL$ uses only acceptance-free step times, so it is known before any request
arrives. At $d{=}7$ and 1M it is $2.7$--$2.9$ for Qwen-35B, $3.6$--$3.9$ for Qwen-122B and
$2.9$--$3.0$ for Nemotron; the ranges come from dense TPOT varying across tasks. The engine already logs the accepted length of every step, so a server can
compare its running $\AL$ with the break-even $\AL$ and lower $d$, or turn speculation off, when
it falls below.

\paragraph{Calibrating the model.}
The break-even test needs only two step times at the deployed depth, from one dense and
one native run at the target context length on any input. Step time is the time per decode iteration, $\AL\times\mathrm{TPOT}$, which any engine
that reports acceptance and throughput provides (in SGLang,
$\text{accept\_len}/\text{throughput}$); $\AL$ cancels, so it does not depend on acceptance.
Measuring native at two or more depths and fitting the line
$\text{step}=c+\gamma\,\tfwd$ adds two things: the break-even $\AL$ at unmeasured depths, so a
server can pick $d$ rather than only switch speculation off, and the split of the tax into
the intercept $c$ and the per-forward slope $\tfwd$ (below). Two refinements
apply. (i)~On Qwen-122B and Nemotron the verify cost jumps once, by ${\approx}6$\,ms, at
$d{=}5$, where its attention crosses a kernel tile boundary; depths on either side need
their own fit (we fit $d{=}5$--$8$; App.~\ref{app:latency}). (ii)~Every term grows with
context, so the constants are calibrated per context length and batch size.
Calibrated on \texttt{niah\_multiquery\_enum} alone, the break-even $\AL$ flags all six
cells of Table~\ref{tab:hero} where native is slower than dense, plus three near-ties where
native is at most $7\%$ ahead. Across $d{=}5$--$8$ on the other inputs, the fitted line
flags all $20$ losing cells (App.~\ref{app:predict}).

\paragraph{Measuring the tax.}
Table~\ref{tab:cost} gives the fitted constants and the measured step at $d{=}7$.
Windowing saves $7.3$--$8.1$\,ms per step, nearly all in the draft phase: $\tverify$ is
unchanged on the Qwen models and drops $1.2$\,ms on Nemotron (Table~\ref{tab:iterphase}).
Both draft terms shrink. The intercept $c$ falls by $5.1$--$6.1$\,ms, almost all of it the
draft's index build, which windowing nearly eliminates. This part depends on the
implementation and a leaner index kernel would shrink it, but windowing stays
$1.07$--$1.16\times$ faster per step even if native's index build were free
(App.~\ref{app:latency}). The slope $\tfwd$ falls $22$--$40\%$, worth $1.9$--$2.9$\,ms at
$\gamma{=}6$. This part comes from reading a bounded window instead of the whole context,
so it holds on any backend, and it matters more as hybrid targets make verification
cheaper. Before windowing, the draft and bookkeeping together cost $92$--$138\%$ of the
bare verify pass, so speculation roughly doubles the step; windowing cuts this to
$45$--$72\%$. The per-step speedup is $1.28\times$ (Nemotron) to $1.44\times$ (Qwen-35B),
and the end-to-end speedup over native is this times the change in acceptance:
\begin{equation}
  \frac{\mathrm{TPOT}^{\text{native}}}{\mathrm{TPOT}^{\text{win}}}
  = \frac{\text{step}^{\text{native}}}{\text{step}^{\text{win}}}
    \cdot \frac{\AL^{\text{win}}}{\AL^{\text{native}}}.
  \label{eq:ratio}
\end{equation}
The first factor is fixed for a model and context length; the second depends on the
workload (\S\ref{sec:mechanism}).

\begin{table}[H]
  \centering
  \caption{Per-step decode cost at 1M, $B{=}1$ (bf16 KV, single B200), native$\to$windowed,
  on \texttt{niah\_multiquery\_enum} (step is within $3\%$ on most inputs and up to $9\%$ shorter on aggregation inputs; App.~\ref{app:predict}). $c$ and
  $\tfwd$ are the line of Eq.~\ref{eq:tpot} fitted over $d{=}5$--$8$ ($R^2{\ge}0.97$;
  Qwen-122B over $d{=}5$--$7$, since $d{=}8$ does not fit at 1M). Step is measured at
  $d{=}7$ ($\gamma{=}6$) without profiling and agrees with $c{+}6\,\tfwd$ within
  $0.3$\,ms. Speedup is $\text{step}^{\mathrm{nat}}/\text{step}^{\mathrm{win}}$, the gain
  at matched acceptance.}
  \label{tab:cost}
  \small
  \setlength{\tabcolsep}{3pt}
  \begin{tabular}{lcccc}
    \toprule
    model & $c$ (ms) & $\tfwd$ (ms) & step (ms) & speedup \\
    \midrule
    Qwen-35B  & 19.3$\to$14.2 & 1.19$\to$0.71 & 26.4$\to$18.3 & \textbf{1.44}$\times$ \\
    Qwen-122B & 27.6$\to$21.5 & 1.16$\to$0.85 & 34.5$\to$26.5 & \textbf{1.30}$\times$ \\
    Nemotron  & 24.6$\to$19.1 & 1.45$\to$1.13 & 33.1$\to$25.8 & \textbf{1.28}$\times$ \\
    \bottomrule
  \end{tabular}
\end{table}

\begin{table}[H]
  \centering
  \caption{Per-iteration wall-clock phases from the nsys trace ($d{=}7$, 1M, $B{=}1$; mean
  over 18 steady iterations, native$\to$windowed). $\tverify^{\mathrm{base}}$: the target's
  verify forward over the $d$-token block; $\tverify^{\mathrm{ovh}}$: speculative bookkeeping (tree build, sampling); together they form $\tverify$. Profiling adds $2$--$5$\,ms per
  iteration relative to Table~\ref{tab:cost}.}
  \label{tab:iterphase}
  \small
  \begin{tabular}{lcccc}
    \toprule
    model & $\tverify^{\mathrm{base}}$ & $\tverify^{\mathrm{ovh}}$ & $\tdraft$ & Iteration \\
          & \multicolumn{4}{c}{\footnotesize (ms/iter, native$\to$windowed)} \\
    \midrule
    q35  & 12.5$\to$12.4 & 4.6$\to$4.6 & \textbf{12.7$\to$4.3} & 29.7$\to$21.3 \\
    q122 & 19.5$\to$19.6 & 4.3$\to$4.2 & \textbf{13.6$\to$7.3} & 37.4$\to$31.1 \\
    nem  & 19.0$\to$19.0 & 5.8$\to$4.6 & \textbf{11.6$\to$4.0} & 36.5$\to$27.6 \\
    \bottomrule
  \end{tabular}
\end{table}

\paragraph{Why native speculation collapses at long context.}
On hard tasks $\AL$ levels off well below $d$, while every extra draft forward still adds
$\tfwd$, which at 1M includes a full-context read. The numerator of Eq.~\ref{eq:tpot}
keeps rising with $d$ and the denominator does not, so TPOT climbs back past dense, which
pays for neither the draft nor the bookkeeping. On the large models the verify tile jump
at $d{=}5$ adds a further ${\approx}6$\,ms that deeper drafts pay without any acceptance
gain. The tax is specific to long context, though not because the draft outgrows the
target: both read $\mathcal{O}(S)$ of KV, so the draft's share of the work is the same at
any length. What changes is whether those reads cost time. A draft layer's KV is
${\approx}1$--$2$\,KB/token, so at $8$--$32$K its $8$--$64$\,MB can fit in the B200's
$126$\,MB L2, potentially reducing the HBM traffic of the $\gamma$ re-reads. At
1M the $1$--$2$\,GB draft KV far exceeds L2, so each draft forward must stream it from HBM. More on-chip capacity
or bandwidth moves the context length at which the tax appears; it does not remove it.

\section{Method: Windowed-MTP}
\label{sec:method}

\paragraph{Mechanism.}
At each draft step, the draft attends only to the first $n_{\text{sink}}$ tokens
(the attention sink) and the most recent $W$ tokens, instead of the whole context
(Fig.~\ref{fig:window}). Each cached key already carries its position, so the draft
sees the same scores for these tokens as before; it simply skips the rest. The target's
verify pass is unchanged and still attends to the full context. The method is
training-free: the engine only changes which cached
entries it hands to the draft (implementation in App.~\ref{app:repro}).

\begin{figure}[!t]
\centering
\adjustbox{max width=\linewidth}{%
\begin{tikzpicture}[x=1pt,y=1pt,font=\footnotesize]
  \node[anchor=east] at (-4,62) {\textbf{Native}};
  \fill[blue!22] (0,52) rectangle (224,72);
  \draw (0,52) rectangle (224,72);
  \node at (112,62) {draft attends \emph{all} $t$ keys (grows to 1M)};
  \node[anchor=east] at (-4,18) {\textbf{Windowed}};
  \fill[blue!22] (0,8) rectangle (26,28); \draw (0,8) rectangle (26,28);
  \node at (13,18) {\scriptsize sink};
  \fill[gray!10] (26,8) rectangle (150,28); \draw[dashed] (26,8) rectangle (150,28);
  \node at (88,18) {\scriptsize dead KV --- reclaimed};
  \fill[blue!22] (150,8) rectangle (224,28); \draw (150,8) rectangle (224,28);
  \node at (187,18) {\scriptsize window $W$};
  \draw[decorate,decoration={brace,amplitude=4pt,mirror}] (0,3) -- (26,3);
  \draw[decorate,decoration={brace,amplitude=4pt,mirror}] (150,3) -- (224,3);
  \node[anchor=north] at (112,-4) {\scriptsize fixed footprint $n_{\text{sink}}{+}W{+}d$ (independent of $t$)};
\end{tikzpicture}%
}
\caption{Draft attention before/after windowing. The \emph{native} MTP draft attends the
full, growing key set, so its per-draft-step cost and KV footprint scale with context
$t$. \emph{Windowed-MTP} restricts the draft to $n_{\text{sink}}$ sink tokens plus the
most recent $W$ keys; the intervening keys are never read and are reclaimed by storing
the draft KV in a compact $n_{\text{sink}}{+}W{+}d$ ring buffer.}
\label{fig:window}
\end{figure}

\begin{widefigure}
  \centering
  \includegraphics[width=\textwidth]{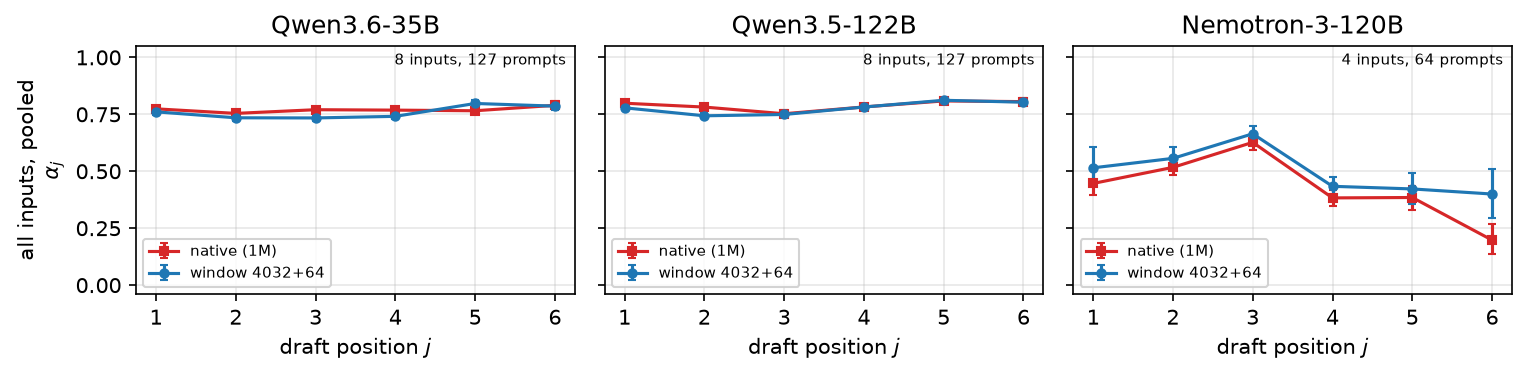}
  \caption{Per-position conditional acceptance $\alpha_j$ ($d{=}7$, $\gamma{=}6$, 1M),
  native (full-1M) vs.\ windowed (4032+64), for the three models (columns), pooled over every
  prompt of every input in Table~\ref{tab:hero} (Qwen: $8$ inputs, $127$ prompts; Nemotron:
  $4$ inputs, $64$ prompts). Error bars: 95\% cluster-bootstrap intervals over prompts,
  stratified by input. Same runs as Table~\ref{tab:hero}.}
  \label{fig:alpha}
\end{widefigure}

\paragraph{Why this preserves the target distribution.}
Let the full-context target induce $p(\cdot \mid x_{<t})$ and the draft induce $q$.
Under speculative sampling, a proposal is accepted under the standard $p/q$ rule; after a
rejection, the replacement is sampled from the residual distribution proportional to
$[p-q]^+$. This procedure returns exactly $p$, regardless of $q$. Under greedy decoding,
the target accepts the matching proposal prefix and appends its own token at the first
mismatch. Replacing a full-attention $q$ with $q^{\text{win}}$ therefore changes only
proposals and acceptance; in exact arithmetic, greedy output is bit-exact.

In practice, \texttt{bf16} SD is not bit-exact against dense. Verifying $d$ tokens
in one batched forward reorders floating-point reductions, so occasional argmax near-ties
flip and compound. This is a property of the verification batch, not of windowing: it
affects native and windowed drafts alike, just as it separates two native depths. The
relevant comparison is therefore windowed versus native, not either versus dense. Our
patch does not modify the target verify path, so the guarantee rests on the standard
verifier; we have not yet quantified the bf16 divergence rates between arms.

\paragraph{Memory.}
Capping the draft's KV pool to a $W+n_{\text{sink}}$ ring buffer frees the remainder of
what would otherwise be a full-length draft cache---an architecturally-fixed
\textbf{7.7--11.1\% of total KV} across our three models (\S\ref{sec:results})---and that
budget converts directly into serving headroom. At 1M on a single B200, with each arm at its best feasible memory setting, the ring buffer lets Qwen-35B ($d{=}7$) seat $B_{\text{eff}}{=}6$ concurrent requests at a $6.28$M-token KV pool where native MTP seats five (Fig.~\ref{fig:pareto}; App.~\ref{app:repro}). The ring keeps
the sink fixed and reuses the recent slots circularly; App.~\ref{app:repro} explains why
it never reads an overwritten entry and needs no rollback after a rejection.

\section{How windowing changes acceptance}
\label{sec:mechanism}

Eq.~\ref{eq:ratio} shows the speedup has two factors: a cost factor
($\text{step}^{\text{native}}/\text{step}^{\text{win}} > 1$, always favorable)
and an acceptance factor $\AL^{\text{win}}/\AL^{\text{native}}$. A naive
expectation is that windowing can only \emph{lose} acceptance (the draft sees
less). In practice the change is usually small and goes both ways: across the $20$
cells the median per-prompt ratio is $0.96$ and is above $1$ on six (Table~\ref{tab:alci}), but it
falls by $16$--$19\%$ (task-level medians) on Qwen-35B NIAH-mv and VT. This section explains why the change is
usually small, and why the cheaper step still outweighs the larger losses.

\paragraph{Window size.}
Fixing $d{=}7$ ($\gamma{=}6$ draft forwards) and sweeping the window $W$ on
\texttt{niah\_\allowbreak multiquery\_\allowbreak enum} at 1M ($16$ prompts, Table~\ref{tab:wsink}), every window
from 1K to 8K beats native on every model: $1.29$--$1.35\times$ on Qwen-35B,
$1.14$--$1.17\times$ on Qwen-122B and $1.27$--$1.31\times$ on Nemotron (median per-prompt
speedup), with every 95\% bootstrap interval above $1.10$. Acceptance is just as flat
(median per-prompt ratio to native $0.93$--$0.96$, $0.88$--$0.90$ and $0.99$--$1.02$). The
needles sit outside both the 4K and 8K windows, so a larger window adds only distant context
and does not raise acceptance; it only re-adds the draft's $\mathcal{O}(S)$ tax. The gain
therefore does not hinge on tuning $W$. On Qwen-35B the default ${\sim}4$K window keeps
$96\%$ of native acceptance and cuts median per-token latency from $5.86$ to $4.17$~ms.

\begin{table}[!htb]
  \centering
  \caption{Window and sink ablation ($d{=}7$, 1M, $B{=}1$, \texttt{niah\_multiquery\_enum},
  $16$ prompts): median $\AL$ and median per-prompt TPOT speedup over the native
  full-context draft. Window columns use sink $64$; sink columns use $W{=}4032$. The
  operating point ($W{=}4032$, sink $64$) is bold.}
  \label{tab:wsink}
  \small
  \setlength{\tabcolsep}{4pt}
  \begin{tabular}{llc|cccc|ccc}
    \toprule
    & & & \multicolumn{4}{c|}{window $W$ (sink $64$)} & \multicolumn{3}{c}{sink ($W{=}4032$)} \\
    model & & native & $960$ & $1984$ & $\mathbf{4032}$ & $8128$ & $0$ & $32$ & $128$ \\
    \midrule
    \multirow{2}{*}{q35}  & $\AL$   & 4.45 & 4.45 & 4.53 & \textbf{4.43} & 4.21 & 4.36 & 4.59 & 4.23 \\
                          & speedup & 1.00$\times$ & 1.32$\times$ & 1.35$\times$ & \textbf{1.34$\times$} & 1.29$\times$ & 1.34$\times$ & 1.39$\times$ & 1.32$\times$ \\
    \multirow{2}{*}{q122} & $\AL$   & 4.76 & 4.30 & 4.25 & \textbf{4.32} & 4.45 & 4.18 & 4.28 & 4.32 \\
                          & speedup & 1.00$\times$ & 1.15$\times$ & 1.14$\times$ & \textbf{1.16$\times$} & 1.17$\times$ & 1.13$\times$ & 1.16$\times$ & 1.17$\times$ \\
    \multirow{2}{*}{nem}  & $\AL$   & 3.63 & 3.70 & 3.70 & \textbf{3.68} & 3.49 & 3.17 & 3.60 & 3.80 \\
                          & speedup & 1.00$\times$ & 1.30$\times$ & 1.31$\times$ & \textbf{1.27$\times$} & 1.27$\times$ & 1.16$\times$ & 1.27$\times$ & 1.28$\times$ \\
    \bottomrule
  \end{tabular}
\end{table}

\paragraph{Sink size.}
The sink costs $64$ of the draft's $4{,}096$ slots, and removing it lowers acceptance on all
three models (Table~\ref{tab:wsink}). The loss is largest on Nemotron: the median per-prompt
ratio to sink $64$ is $0.87$, which cuts its speedup over native from $1.27\times$ to
$1.16\times$. Qwen-122B follows ($0.95$, $1.16{\to}1.13\times$), then Qwen-35B ($0.99$); with
$16$ prompts no interval excludes $1$. A $32$-token sink recovers the loss on all three
models, and from $32$ to $128$ the median ratio to sink $64$ stays within $0.02$, so the sink
size needs no tuning either.

\paragraph{Across inputs.}
Across the full $d{=}7$ sweep
(Table~\ref{tab:hero}) the acceptance factor varies in sign with input and model, but in no
model--task cell does it cancel the per-step saving: the median nat/win is ${\ge}1.14\times$
on every cell. A direct in-run probe (Table~\ref{tab:top1}) shows why the change is
usually small: windowing leaves the draft's top-1 proposal unchanged $86$--$94\%$ of the
time.

\paragraph{Per-position decomposition.}
We read SGLang's per-request acceptance histogram $H[n]$ (number of verify steps
that accepted exactly $n$ draft tokens) and derive the \emph{conditional}
acceptance rate at draft position $j$,
\begin{equation}
\begin{aligned}
  \alpha_j &= \Pr[\text{accept } j \mid \text{reached } j]
            = \frac{\sum_{n\ge j} H[n]}{\sum_{n\ge j-1} H[n]}, \\
  \AL &= 1 + \sum_{j\ge 1}\ \textstyle\prod_{k\le j}\alpha_k .
\end{aligned}
  \label{eq:alpha}
\end{equation}
Figure~\ref{fig:alpha} shows $\alpha_j$ ($d{=}7$, $\gamma{=}6$, 1M) for native vs.\
windowed drafts, pooled over every prompt of every input in Table~\ref{tab:hero}. The windowed
profile \emph{tracks} the native one position by position. On both Qwen models the two differ
by at most $0.04$ at any position (largest: Qwen-122B at $j{=}2$, $0.78{\to}0.74$), and
$\alpha_1$ (local prediction) by at most $0.02$. With $127$ prompts per model some of these
gaps are statistically resolvable, but none is large. On Nemotron the windowed draft accepts
more at every position. Neither profile collapses at the deep, compounding positions. Single inputs spread more widely. The largest per-input drop is
Qwen-35B NIAH-mv at $j{=}2$ ($-0.18$), the cell whose median $\AL$ falls most, and Qwen-122B, the strongest drafter, loses $11$--$13\%$ on multi-query and multi-value retrieval (median ratios $0.89$ and $0.87$, Table~\ref{tab:alci}), where far context is \emph{informative} rather than dilutive. Across all inputs
(Table~\ref{tab:hero}) the change is usually small, but not always: Qwen-35B's median loss is $19\%$
on NIAH-mv and $16\%$ on VT, giving end-to-end gains of $1.15\times$ and $1.21\times$.

Pooled over inputs, the acceptance difference arises mainly past the first draft
position: $\alpha_1$ barely moves on the Qwen models, so a shallow draft is nearly
windowing-invariant and the gap emerges only as $d$ grows (App.~\ref{app:scope}). The
cost factor is favorable at every depth; acceptance is a second-order term on top.

\paragraph{Why windowing rarely changes the draft's choice.}
All three drafts sit on recurrent-hybrid models (Gated-DeltaNet for Qwen, Mamba2 for
Nemotron), whose recurrent layers already carry long-range information, so the draft's
reread of distant context mostly repeats what it already has. This holds even for
Nemotron, whose NoPE attention puts ${\approx}84\%$ of its softmax mass outside the
window, yet whose top-1 proposal is preserved $86\%$ of the time. On the RoPE Qwen
drafts, the ${\approx}20$--$25\%$ of mass beyond the window is diffuse rather than peaked
on a few informative keys (App.~\ref{app:mech}, Table~\ref{tab:attnmass}). The argument
is specific to hybrid targets; a draft on a dense-attention model may lean more on distant
context. We treat it as interpretation, not proof.

\section{Experiments}
\label{sec:results}

\paragraph{Setup.}
Unless noted, we use one NVIDIA B200 (TP1), bf16 KV, and SGLang with its built-in
MTP speculative decoding and CUDA graphs; Fig.~\ref{fig:tp} instead uses two B200s at TP2. The models
are Qwen3.6-35B-A3B and Qwen3.5-122B-A10B (GDN-MoE, MoE-FFN MTP), and
Nemotron-3-Super-120B-A12B (Mamba2-hybrid, eight GQA full-attention layers, MoE-FFN
MTP, NoPE); checkpoints are in App.~\ref{app:repro}. Windowed-MTP uses
$W{=}4032$, $n_{\text{sink}}{=}64$.

At a fixed 1.04M-token input, we evaluate RULER~\citep{hsieh2024ruler} retrieval and
aggregation tasks, LongBench-v2~\citep{bai2024longbench2} code-reasoning QA, and
BABILong~\citep{kuratov2024babilong} multi-fact QA. We report acceptance, time per output token (TPOT), and speedup over native MTP and dense decoding.  Unless noted, all arms decode greedily (temperature $0$)
with a repetition penalty of $1.15$, which prevents enumeration loops from artificially
inflating acceptance. Each task
uses QA mode, stopping at EOS or 512 new tokens. Thus every timed decode step attends to
an effectively fixed ${\sim}1$M-token cache---the long-prompt, short-completion regime.
The same draft-cost mechanism applies when a short prompt reaches that length through
generation instead.

No input is tiled or repeated to reach length, all inputs come from seeded
generators that reproduce byte-for-byte, and positional validity at 1M is
handled at the model level (YaRN for Qwen, native NoPE for Nemotron), applied identically
to the dense baseline and every speculative variant (App.~\ref{app:repro}).
Table~\ref{tab:hero} reports multi-prompt results: $16$ independently generated prompts
per task ($15$ for Code-QA); each model--task cell reports the median over them. The cost model, the
depth ablation and the batch sweeps use one reference prompt per task
(the first of the $16$) unless stated otherwise; the sampling and online runs are
described where they appear.

\paragraph{Results across all inputs.}
Table~\ref{tab:hero} reports every task at $d{=}7$. Windowed-MTP is faster than native
MTP on every model--task cell, by a median $1.14$--$1.56\times$, and $1.03$--$2.54\times$ faster than
dense decoding. Native MTP, in contrast, is slower than dense on six cells and ties it on a
seventh.
Acceptance changes little: the median windowed-to-native $\AL$ ratio is $0.96$, $16$ of
the $20$ cells are within $\pm15\%$ (unrounded), and windowing raises $\AL$ on six (Table~\ref{tab:alci}).
Eq.~\ref{eq:ratio} splits each prompt's speedup into a per-step factor and an acceptance
factor. The per-step factor is nearly constant across prompts (cell medians
$1.40$--$1.44\times$ on Qwen-35B, $1.28$--$1.31\times$ on Qwen-122B and $1.26$--$1.27\times$
on Nemotron), so variation across prompts and tasks tracks
acceptance, and the smallest gains come where acceptance drops most: Qwen-35B NIAH-mv ($\AL$ ratio $0.81$, $1.15\times$) and Qwen-122B NIAH-mv and
BABILong ($0.87$ and $0.90$, both $1.14\times$). Even there, the cheaper step outweighs
the lost acceptance (per-prompt spread in App.~\ref{app:tables}). The per-step gain is smallest on
Nemotron, whose windowable share of draft cost is smallest (\S\ref{sec:discussion}).

\begin{widetable}
  \centering
  \caption{All inputs at 1M, $d{=}7$ ($\gamma{=}6$), $B{=}1$, greedy (temperature $0$). TPOT in ms; spd is
  speedup over dense (no speculation); nat/win is native TPOT over windowed TPOT (speedup over native).
  All arms run the same $16$ prompts per task ($15$ for Code-QA); $\AL$ and TPOT are medians
  over prompts, and spd and nat/win are medians of per-prompt ratios (95\% intervals in
  App.~\ref{app:tables}). Nemotron is shown on its four
  long-decode tasks; on NIAH-s, VT, Code-QA and BABILong it answers in $4$--$12$ tokens,
  too briefly to time.}
  \label{tab:hero}
  \small
  \setlength{\tabcolsep}{4.5pt}
  \begin{tabular}{llcccccccc}
    \toprule
    & & Dense & \multicolumn{3}{c}{native MTP ($d{=}7$)} & \multicolumn{3}{c}{Windowed-MTP ($d{=}7$)} & nat/win \\
    \cmidrule(lr){4-6}\cmidrule(lr){7-9}\cmidrule(lr){10-10}
    Model & Input & TPOT & $\AL$ & TPOT & spd & $\AL$ & TPOT & spd & (TPOT) \\
    \midrule
\multirow{8}{*}{Qwen3.6-35B}
  & NIAH-s & 9.84 & 5.07 & 5.19 & 1.88$\times$ & 4.72 & \textbf{3.90} & 2.54$\times$ & \textbf{1.35$\times$} \\
  & NIAH-mv & 9.84 & 5.42 & 4.80 & 2.05$\times$ & 4.39 & \textbf{4.24} & 2.32$\times$ & \textbf{1.15$\times$} \\
  & NIAH-mq & 9.79 & 4.45 & 5.86 & 1.67$\times$ & 4.43 & \textbf{4.17} & 2.35$\times$ & \textbf{1.34$\times$} \\
  & VT & 9.72 & 4.18 & 6.36 & 1.53$\times$ & 3.63 & \textbf{5.10} & 1.90$\times$ & \textbf{1.21$\times$} \\
  & CWE & 9.21 & 2.74 & 8.83 & 1.05$\times$ & 2.81 & \textbf{6.15} & 1.51$\times$ & \textbf{1.45$\times$} \\
  & FWE & 9.25 & 2.85 & 8.68 & 1.07$\times$ & 3.02 & \textbf{5.73} & 1.61$\times$ & \textbf{1.56$\times$} \\
  & Code-QA & 9.96 & 3.30 & 8.05 & 1.24$\times$ & 3.14 & \textbf{5.98} & 1.67$\times$ & \textbf{1.37$\times$} \\
  & BABILong & 9.65 & 3.07 & 8.38 & 1.17$\times$ & 2.78 & \textbf{6.54} & 1.48$\times$ & \textbf{1.28$\times$} \\
  \midrule
\multirow{8}{*}{Qwen3.5-122B}
  & NIAH-s & 9.54 & 4.04 & 8.62 & 1.11$\times$ & 3.75 & \textbf{7.17} & 1.33$\times$ & \textbf{1.21$\times$} \\
  & NIAH-mv & 9.54 & 4.81 & 7.16 & 1.33$\times$ & 4.34 & \textbf{6.07} & 1.57$\times$ & \textbf{1.14$\times$} \\
  & NIAH-mq & 9.56 & 4.76 & 7.20 & 1.33$\times$ & 4.32 & \textbf{6.04} & 1.58$\times$ & \textbf{1.16$\times$} \\
  & VT & 9.60 & 3.82 & 9.15 & 1.05$\times$ & 3.67 & \textbf{7.30} & 1.31$\times$ & \textbf{1.25$\times$} \\
  & CWE & 8.86 & 3.28 & 9.61 & 0.92$\times$ & 3.27 & \textbf{7.47} & 1.19$\times$ & \textbf{1.27$\times$} \\
  & FWE & 8.77 & 3.45 & 9.17 & 0.96$\times$ & 3.44 & \textbf{7.12} & 1.24$\times$ & \textbf{1.34$\times$} \\
  & Code-QA & 9.57 & 3.16 & 11.07 & 0.86$\times$ & 3.32 & \textbf{8.01} & 1.19$\times$ & \textbf{1.31$\times$} \\
  & BABILong & 9.32 & 3.64 & 9.33 & 1.00$\times$ & 3.08 & \textbf{8.54} & 1.10$\times$ & \textbf{1.14$\times$} \\
  \midrule
\multirow{4}{*}{Nemotron-3-120B}
  & NIAH-mv & 11.35 & 2.86 & 11.66 & 0.97$\times$ & 3.02 & \textbf{8.70} & 1.30$\times$ & \textbf{1.26$\times$} \\
  & NIAH-mq & 11.39 & 3.63 & 9.09 & 1.25$\times$ & 3.68 & \textbf{6.97} & 1.63$\times$ & \textbf{1.27$\times$} \\
  & CWE & 10.90 & 1.64 & 18.68 & 0.58$\times$ & 2.47 & \textbf{9.83} & 1.11$\times$ & \textbf{1.47$\times$} \\
  & FWE & 10.92 & 1.64 & 18.51 & 0.59$\times$ & 2.30 & \textbf{10.76} & 1.03$\times$ & \textbf{1.53$\times$} \\
    \bottomrule
  \end{tabular}
\end{widetable}

\begin{figure}[!t]
  \centering
  \includegraphics[width=0.6\textwidth]{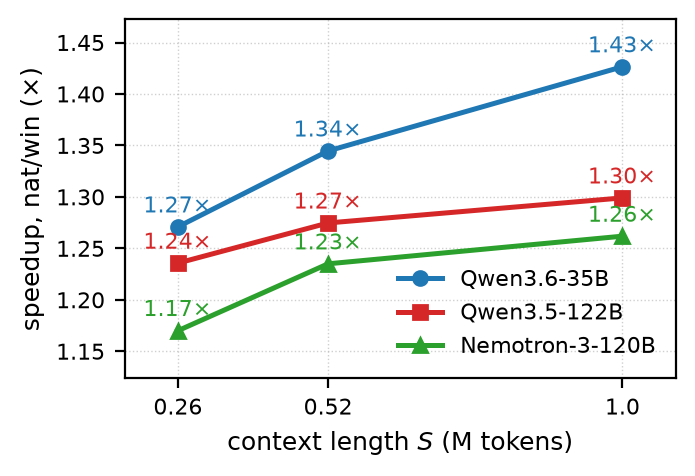}
  \caption{The windowing wedge \emph{widens with context}. Matched-acceptance
  speedup $\text{step}^{\mathrm{nat}}/\text{step}^{\mathrm{win}}$
  ($T_{\text{iter}}{=}\AL\cdot\text{TPOT}$, so acceptance cancels) vs.\ context length
  $S$, at $d{=}7$, $B{=}1$, bf16, on \texttt{niah\_multiquery\_enum}. The 1M endpoints
  match Table~\ref{tab:cost} within run-to-run noise (${\le}0.03\times$).}
  \label{fig:ctxscale}
\end{figure}

\paragraph{The per-step cost saving grows with context.}
The windowing wedge widens as context grows because windowing removes the \emph{draft's}
$\mathcal{O}(S)$ context cost: as $S$ grows the native draft's full-context scan adds increasing
cost, while the windowed draft's attention working set stays fixed (4K is $1.6\%$ of 261K but $0.4\%$ of
1M); target verification remains context-dependent under both. Sweeping the true-1M input (\texttt{niah\_multiquery\_enum}, $d{=}7$, matched
acceptance, Table~\ref{tab:cost}) from 261K to 1M (Fig.~\ref{fig:ctxscale}), the
matched-acceptance speedup rises for all three models (Qwen-35B
$1.27{\to}1.43\times$, Qwen-122B $1.24{\to}1.30\times$, Nemotron $1.17{\to}1.26\times$). The absolute per-step saving $\Delta$ roughly doubles in each case
($3.4{\to}7.9$, $3.5{\to}7.8$, $3.2{\to}6.8$\,ms).

\paragraph{Best-depth baseline.}
A practitioner tunes draft depth per model and keeps the fastest, so the fair comparison
is native at \emph{its} best depth vs.\ windowed at \emph{its} best depth.
Table~\ref{tab:bestdepth} sweeps every depth from $d{=}2$ to $8$ on the multi-query
retrieval input at 1M (Qwen-122B stops at $d{=}7$, as $d{=}8$ does not fit its KV pool). \textbf{At each method's best depth, Windowed-MTP is
faster than native on all three models} ($1.18$--$1.41\times$), and it is also faster at
every individual depth. The best depths differ. Native's TPOT is non-monotonic in $d$:
each extra draft forward pays the full-context read, and the large models also pay the
verify jump at $d{=}5$, so Nemotron's native optimum is $d{=}4$. The windowed draft is
cheap, so its optimum is set by acceptance instead: it sits deeper than native on
Qwen-35B ($d{=}7$ vs.\ $6$) and shallower on Qwen-122B ($d{=}4$ vs.\ $7$), where the
cheaper step more than pays for the tokens it gives up.

\begin{table}[!t]
  \centering
  \caption{Best-depth baseline (1M, $B{=}1$, \texttt{niah\_\allowbreak multiquery\_\allowbreak enum}): native
  vs.\ windowed, each at its own min-TPOT depth over $d{=}2$--$8$ (Qwen-122B: $d{\le}7$,
  as $d{=}8$ exceeds the KV pool for both arms). TPOT in ms, best $d$ in parens; dense
  TPOT for reference. Single run per depth.}
  \label{tab:bestdepth}
  \small
  \setlength{\tabcolsep}{4pt}
  \begin{tabular}{lcccc}
    \toprule
    model & dense & native (d) & windowed (d) & nat/win \\
    \midrule
    Qwen-35B  & 9.90  & 5.47 (6) & 3.89 (7) & \textbf{1.41}$\times$ \\
    Qwen-122B & 9.56  & 6.09 (7) & 5.15 (4) & \textbf{1.18}$\times$ \\
    Nemotron  & 11.27 & 7.42 (4) & 5.34 (4) & \textbf{1.39}$\times$ \\
    \bottomrule
  \end{tabular}
\end{table}

\begin{widefigure}
  \centering
  \includegraphics[width=0.75\textwidth]{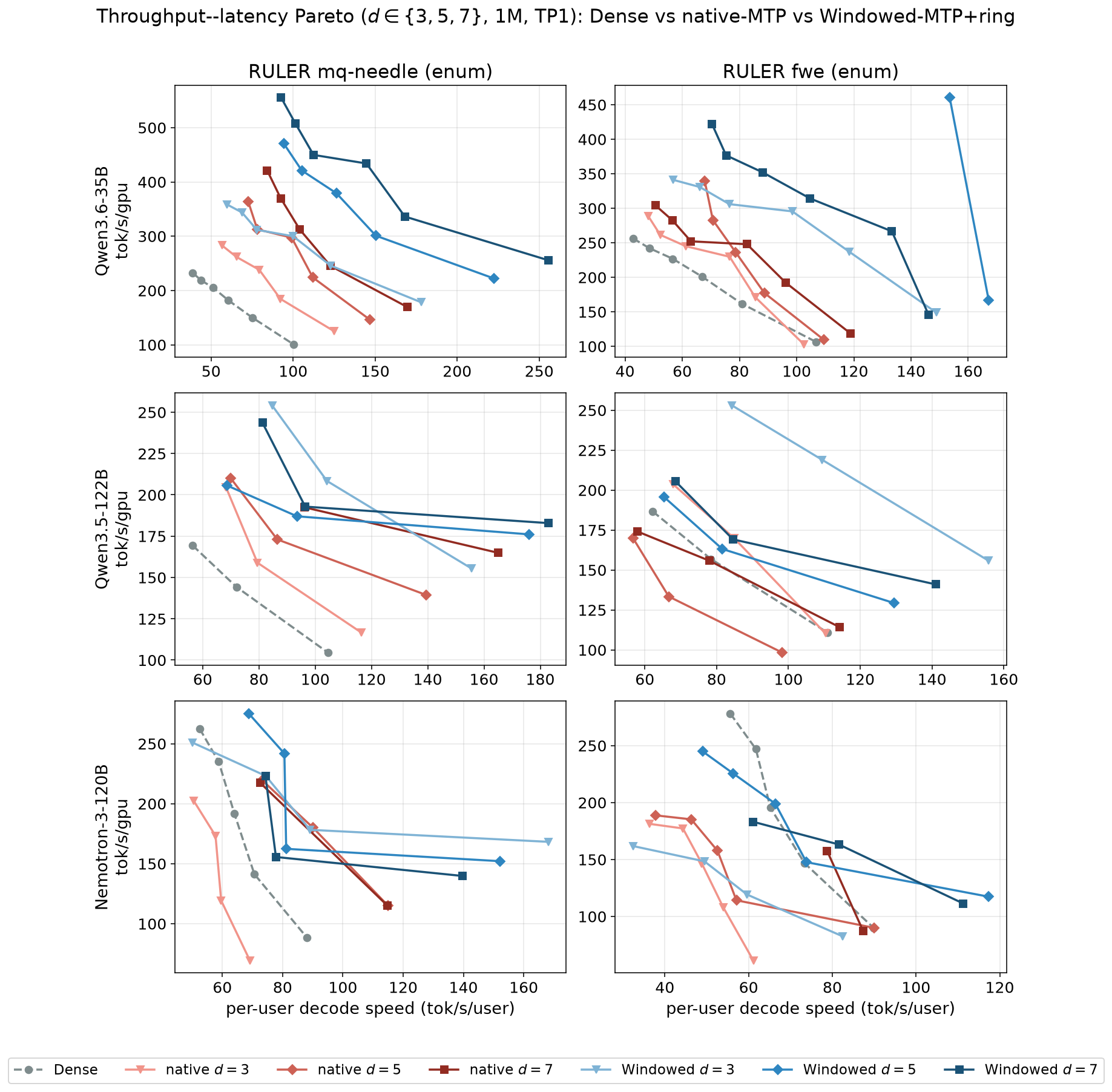}
  \caption{Throughput--latency Pareto at 1M ($B$ sweep over $d\in\{3,5,7\}$; single B200,
  TP1). Rows: Qwen3.6-35B, Qwen3.5-122B, Nemotron-3-120B; columns:
  \texttt{niah\_multiquery\_enum} and \texttt{fwe} (up-and-right is better).
  Windowed-MTP+ring (blue) holds the frontier over native MTP (red) and Dense (grey) in
  five of six panels, the exception being Nemotron+FWE (see text), and also fits more
  resident 1M requests (one more on 35B at each arm's best feasible memory setting).}
  \label{fig:pareto}
\end{widefigure}

\paragraph{Sampling at temperature 1.}
Our main runs are greedy. To check that the gain carries over to sampling, we rerun
Qwen-35B at 1M and $d{=}7$ with temperature $1$ and the model's released default
sampling settings (top-$k$ $20$, top-$p$ $0.95$, from its \texttt{generation\_config}), on every prompt of three inputs, averaging three seeds per prompt (Table~\ref{tab:temp1}).
Sampling lowers native acceptance most on Code-QA (median $3.30{\to}2.77$, greedy vs.\
sampled), while the windowed-to-native $\AL$ ratio stays near $1$ ($0.96$--$1.07$). The
per-step speedup is $1.39$--$1.44\times$ on every seed, matching the greedy $1.44\times$
(Table~\ref{tab:cost}), so windowing is $1.39$--$1.48\times$ faster than native. On CWE and
Code-QA, native is no faster than dense ($0.95\times$ and $1.01\times$, slower on $15$ of
$16$ and $7$ of $15$ prompts), and windowing brings speculation back above it ($1.40\times$
and $1.36\times$).
\begin{table}[!t]
  \centering
  \caption{Temperature-1 sampling (top-$k$ $20$, top-$p$ $0.95$), Qwen-35B, 1M, $d{=}7$,
  $B{=}1$; medians over $16$ prompts ($15$ for Code-QA) of per-prompt means over seeds
  $0$--$2$. TPOT in ms (dense/native/windowed); nat/win is the median per-prompt native over
  windowed TPOT.}
  \label{tab:temp1}
  \small
  \setlength{\tabcolsep}{3.5pt}
  \begin{tabular}{lcccc}
    \toprule
    input & $\AL^{\mathrm{nat}}$ & $\AL^{\mathrm{win}}$ & TPOT (d/n/w) & nat/win \\
    \midrule
    NIAH-mq & 4.57 & 4.36 & 9.94/5.89/4.33  & \textbf{1.39}$\times$ \\
    CWE     & 2.53 & 2.68 & 9.35/9.84/6.66  & \textbf{1.48}$\times$ \\
    Code-QA & 2.77 & 2.59 & 10.07/9.94/7.38 & \textbf{1.43}$\times$ \\
    \bottomrule
  \end{tabular}
\end{table}

\paragraph{Memory.}
From SGLang's KV-pool allocation logs at 1M, the draft (MTP) KV pool is a
\textbf{measured 7.7--11.1\%} of total KV ($7.7$/$9.1$/$11.1\%$ for
Qwen-122B/Qwen-35B/Nemotron) --- matching the architectural $1/(F{+}1)$ share
($F$ full-attention layers ${=}12$/$10$/$8$ of $48$/$40$/$88$). Both pools are sized
up front to the startup KV budget, so this share is a property of the architecture and
that budget rather than of any request: invariant to GPU size and KV precision, with
the absolute GB scaling linearly (App.~\ref{app:repro}). Windowing reads only
$\min(\text{seq}, W{+}n_{\text{sink}})$ base tokens per draft step (structurally
guaranteed by the draft-KV index builder), so at 1M ($W{+}n_{\text{sink}}{=}4096$,
${\approx}0.4\%$ of a 1.04M-token request) \textbf{$>99\%$ of the draft pool is dead and reclaimable}.
Every Windowed-MTP result here runs the ring ($n_{\text{sink}}{+}W{+}d$ slots per
request, confirmed in the same logs). Its payoff is the batch headroom in
Fig.~\ref{fig:pareto}: the windowed arm seats one
more concurrent 1M request, while native's full-length draft pool crowds the target KV
and runs out of KV capacity a batch earlier. This holds at each arm's best feasible memory fraction; with windowing given less memory than native ($0.90$ vs.\ $0.92$), both seat five (App.~\ref{app:repro}).

\begin{widefigure}
  \centering
  \includegraphics[width=\textwidth]{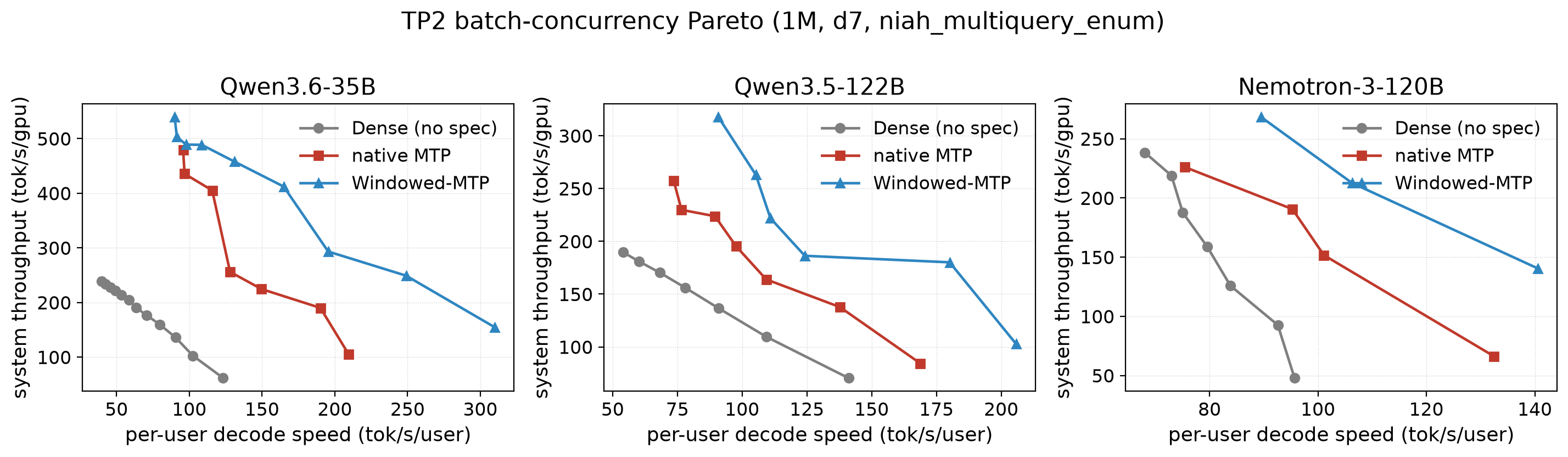}
  \caption{TP2 batch-concurrency Pareto \emph{frontier} (2$\times$B200, 1M, $d{=}7$,
  \texttt{niah\_multiquery\_enum}): per-user decode speed (x)
  vs.\ per-physical-GPU throughput (y), sweeping batch $B$ (non-dominated points only).
  Windowed-MTP (blue) vs.\ native MTP (red) vs.\ Dense (grey).}
  \label{fig:tp}
\end{widefigure}

\paragraph{Throughput--latency Pareto.}
Figure~\ref{fig:pareto} sweeps batch size $B$ at 1M across all three models (rows) and
both workloads (columns) on a single B200, over $d\in\{3,5,7\}$. Taking Qwen-35B (top
row), Windowed-MTP \emph{Pareto-dominates} both Dense and native MTP on both axes at
every batch: at $B{=}1$ it delivers $2.5\times$ the per-user decode speed of Dense (255.6 vs.\ 100.4 tok/s/user), and its peak system throughput (556 tok/s/GPU at $B{=}6$) is
$2.4\times$ Dense's and $1.3\times$ native's. Dominance holds in five of six panels; the
exception is Nemotron+FWE, our lowest-acceptance workload, where the frontier
splits---windowing wins the latency end but Dense wins peak throughput at high batch, the
expected boundary where low acceptance makes speculation unprofitable once the batch is
compute-bound.

\paragraph{Tensor-parallel scaling and batch concurrency (TP2).}
TP2's \emph{primary} purpose here is \emph{capacity}: two B200s hold more concurrent 1M
requests than one. We sweep $B$ at TP2 (2$\times$B200, 1M, $d{=}7$,
\texttt{niah\_multiquery\_enum}) with the radix cache disabled (as for every batch sweep,
so each seated request carries a full, independent 1M KV), from $B{=}1$ to the largest
batch seating a full 1M KV per request ($B{=}12$ for Qwen-35B, $B{=}7$ otherwise).
\textbf{Windowed-MTP's frontier Pareto-dominates both Dense and native MTP for all three
models} (Fig.~\ref{fig:tp}): at equal batch the cheaper step raises the throughput ceiling $1.13$--$1.23\times$ above native and
$1.1$--$2.3\times$ above Dense. Points use measured TPOT, which includes acceptance. We do not compare
$\AL$ across TP degree, as it is not bit-stable there (head splitting reorders the \texttt{bf16} reductions, \S\ref{sec:method}).
Nemotron's frontier varies more run-to-run but stays windowed-dominated.

\paragraph{Online serving under continuous arrivals.}
Every result so far fixes the batch, so we also test under a real scheduler: Qwen-35B at
TP2, Poisson arrivals at an offered rate of $\lambda$ requests/s over mixed 131K--1.04M
contexts, the prefix cache on, and $4096$-token generations, with all three arms
replaying the identical trace at a matched memory fraction (App.~\ref{app:online}). At
$\lambda\in\{0.02,0.05\}$, Windowed-MTP's paired median speedup is $1.15$--$1.21\times$
over native and $2.0$--$2.2\times$ over dense (Table~\ref{tab:onlinepaired}), and it meets
a $25$\,ms p95-TPOT target at $0.05$~req/s, where dense and native meet it only at $0.02$
(Table~\ref{tab:onlineslo}). The gain needs long generations, and it does not raise peak
throughput: from $\lambda{=}0.075$ up, all arms plateau near $0.06$~req/s, because at 1M the
server is bound by prefill, which windowing does not touch (App.~\ref{app:online}).

\paragraph{Cross-hardware generality (H100).}
The win is not specific to Blackwell memory bandwidth. Reproducing the single-GPU result
on one NVIDIA H100-80GB with the \emph{public} FP8 checkpoint
(\texttt{RedHatAI/Qwen3.6-35B-A3B-FP8}; $B{=}1$, 1M, $d{=}7$, \texttt{bf16} KV, same
headline input), Windowed-MTP is again faster than Dense ($2.43\times$) and native
($1.48\times$) and keeps acceptance ($4.34{\to}4.53$; Table~\ref{tab:h100}).

\begin{table}[t]
  \centering
  \caption{H100-80GB replication: Qwen-35B \emph{FP8} (public checkpoint), 1M, $B{=}1$,
  $d{=}7$, \texttt{bf16} KV, \texttt{niah\_multiquery\_enum}. Windowed-MTP is faster than
  Dense and native.}
  \label{tab:h100}
  \small
  \begin{tabular}{lccc}
    \toprule
    method & $\AL$ & TPOT (ms) & speedup \\
    \midrule
    Dense       & 1.00 & 12.26 & $1.00\times$ \\
    native      & 4.34 & 7.48  & $1.64\times$ \\
    windowed    & 4.53 & 5.05  & $\mathbf{2.43\times}$ \\
    \bottomrule
  \end{tabular}
\end{table}

\paragraph{The backend is not the confound.}
Native runs on FlashInfer, but we built the ring on Triton because its kernels are
simpler to modify for the slot remap. Triton is if anything a \emph{handicap}: on Qwen-35B
CWE at $d{=}5$, its draft-extend kernel makes native's TPOT ${\sim}6\times$ worse than on FlashInfer ($8.13\!\to\!50.59$\,ms), and its cost grows with accept
length. Even so, windowed-on-Triton beats native-on-FlashInfer ($5.37$\,ms). A FlashInfer implementation of the ring remains unmeasured
(App.~\ref{app:scope}).

\section{Discussion and limitations}
\label{sec:discussion}

\paragraph{Choice of baseline.}
We compare against the \emph{native full-context MTP} head the target already ships,
with its window disabled and weights, verifier and tree shape fixed. Acceptance-optimizing
drafts (EAGLE-style trees~\citep{li2024eagle2}, EAGLE-3~\citep{li2025eagle3}, DeepSeek's
speculative stack~\citep{deepspec2025}) are complementary: comparing against them would
conflate draft training quality with attention cost, and an EAGLE-3 head attends over the
full KV cache, so it pays the same $\mathcal{O}(S)$ tax and the same window applies.
Block-diffusion drafters such as DFlash~\citep{dflash2026} keep full-context KV in each
draft layer too; measuring windowing on them is future work.

\paragraph{Why not dynamic KV selection?}
A dynamic selector (per-step top-$k$ as in Quest~\citep{tang2024quest}, or SpecExtend's
cross-model retrieval~\citep{cha2026specextend}) must keep the \emph{full} KV resident to
select from, forfeiting the draft-pool reclamation that is a core benefit here, and its
scoring and irregular gather add compute to the very forward we are cheapening---while a
static sink+window already keeps acceptance close to native on most inputs
(\S\ref{sec:mechanism}).

\paragraph{When to window, and how to pick $W$.}
Windowing pays off when the draft's per-step KV cache read is a non-trivial share of the
draft forward, i.e.\ at long context. When $S\lesssim W$ the window already spans the
whole sequence, so Windowed-MTP \emph{degenerates to native MTP} at no cost: there is
never a correctness reason to disable it, and a server can gate it on a context-length
threshold. For the remaining knob $W$, at 1M and $d{=}7$ every window from 1K to 8K is
net-positive on all three models (median $1.14$--$1.35\times$, Table~\ref{tab:wsink}), and
each model's speedup varies by at most $0.06$ across windows, so $W$ needs no per-input
tuning. We use $W{=}4032$, $n_{\text{sink}}{=}64$ everywhere.

\paragraph{Adaptive draft depth.}
Because windowing changes the context-dependent part of the draft cost from $\propto d\cdot S$ to $\propto d\cdot W$ (\S\ref{sec:motivation}), the best draft depth is set mostly by
acceptance rather than by a rising draft tax: the windowed optimum is $d{=}7$ on Qwen-35B but
$d{=}4$ on the other two models (Table~\ref{tab:bestdepth}). An \emph{adaptive}-$d$ controller is therefore
appealing and cheap under windowing, and we leave it to future work. Our headline numbers
fix $d{=}7$ to contrast native and windowed at the \emph{same} depth; windowing also wins
on every model when each method uses its own best depth.

\paragraph{Composition, precision, and prefill.}
Three scope questions have short answers, developed in App.~\ref{app:scope}. Windowing
changes which KV the draft reads, not the shape of the candidate set, so it composes with
tree speculation unchanged, needing only a wider ring
($n_{\text{sink}}{+}W{+}d{\cdot}k$ for width $k$). Under fp8 KV the per-step speedup narrows
to $1.14$--$1.26\times$ at 261K and 1M but persists, since windowing cuts the \emph{number}
of entries read while the dtype sets bytes per entry. Finally, prefill
leaves a further gain untapped. During prefill the
draft runs its layer over all $S$ prompt tokens and writes a key/value entry for
each, but once decoding starts the window reads only the sink and the last $W$ of them;
the rest are written once and never read. The draft could therefore skip those positions
during prefill, turning its $O(S)$ draft prefill into an $O(W{+}n_{\text{sink}})$ one and
cutting TTFT. Our gains are decode-side only, and that kernel work---the one lever that would also lift
the prefill-bound ceiling of App.~\ref{app:online}---is future work.

\paragraph{Limitations.}
(1) The size of the gain depends on the draft and the
framework. Windowing saves only the draft's attention read, so a draft whose step is
dominated by other work gains less. All three MTP heads have 2 KV heads, but Nemotron's
are half as wide ($\text{head\_dim}{=}128$ vs.\ $256$), so its draft reads half the KV
bytes per token, and its MoE FFN is heavier (top-22 of 512 experts vs.\ top-8 of 256);
its gain is $1.28\times$ versus $1.30$--$1.44\times$ for the Qwen drafts
(Table~\ref{tab:cost}). Part of the saving also comes from the draft's full-context
KV-index build, which windowing removes along with the reads; an engine with a cheaper
index build would see a smaller gain, but windowing would still be $1.07$--$1.16\times$
faster per step even if that build cost nothing (App.~\ref{app:latency}). (2) Headline numbers are $B{=}1$, TP1, single-GPU medians over $15$--$16$ prompts per task, one run each. Acceptance in \S\ref{sec:mechanism} pools all prompts; the cost (nearly constant across prompts), top-1 and attention-mass analyses use one. The advantage is not confined to $B{=}1$: a batch sweep to $B{=}6$
(Fig.~\ref{fig:pareto}), a TP2 sweep to $B{=}12$ (Fig.~\ref{fig:tp}), and a
continuous-arrival run (App.~\ref{app:online}) all show
windowing ahead. (3) Distribution preservation is exact only in exact arithmetic: \texttt{bf16} batched verification is not bit-exact against dense for either draft (\S\ref{sec:method}).

\section{Conclusion}
Beside a hybrid target at 1M, a shipped full-attention MTP head's KV reads are a large
share of the decode step and can make deep speculation slower than none. A training-free, serving-time window
on draft attention (Windowed-MTP) removes most of that working set. At 1M it delivers a
nearly input-invariant $1.28$--$1.44\times$ acceptance-free per-decode-step speedup
on three models from two hybrid families, and the gap
grows with context. Acceptance falls on most tasks, by at most $19\%$ per task,
but the cheaper step outweighs it: at the default window the paired median
favors windowing on every model--task cell at $B{=}1$. A ring buffer turns the smaller
working set into physical capacity: $7.7$--$11.1\%$ of total KV at 1M.

\bibliography{refs}

\appendix

\section{Experimental details}
\label{app:repro}

\paragraph{How the 1M inputs are built.} We never tile or repeat text. RULER inputs follow
RULER's construction (needles and queries hidden in a distractor haystack of controlled
length). LongBench-v2 and BABILong use real documents longer than 1M tokens, cut from the
end to length. Every input comes from a seeded generator and is byte-identical across runs.

\paragraph{Models (exact checkpoints).} \texttt{RedHatAI/\allowbreak Qwen3.6-\allowbreak 35B-\allowbreak A3B-\allowbreak NVFP4} and
\texttt{nvidia/\allowbreak Qwen3.5-\allowbreak 122B-\allowbreak A10B-\allowbreak NVFP4} (NVFP4 weights, with a YaRN
\texttt{rope\_scaling} factor of 4 applied to the \texttt{config.json} to reach 1M
from the native 262{,}144-token window) and
\texttt{nvidia/\allowbreak NVIDIA-\allowbreak Nemotron-\allowbreak 3-\allowbreak Super-\allowbreak 120B-\allowbreak A12B-\allowbreak NVFP4} (NVFP4, native NoPE
attention, so no RoPE change). All three have quantized weights. The checkpoints ship FP8
KV scales, which we override so that every run stores KV in bf16
(\texttt{kv\_cache\_dtype=bfloat16}).

\paragraph{Engine and flags.} We serve with SGLang $0.5.12$.post1 (commit
\texttt{3489ce04}) plus our patch (Windowed-MTP and the draft-KV ring buffer; \url{\artifacturl}).
The windowing is merged upstream as SGLang
\href{https://github.com/sgl-project/sglang/pull/32673}{PR~\#32673}
(\texttt{-{}-speculative-draft-window-size}, \texttt{-{}-speculative-draft-sink-size});
the ring buffer is not yet upstream. All B200 and H100 runs use one image (NGC
\texttt{sglang:26.06-py3}: FlashInfer $0.6.12$, Triton $3.6.0$, PyTorch $2.13.0$a0, CUDA $13.3$). \texttt{mem-fraction-static} is $0.85$ for single-request runs. In the batch and
Pareto sweeps each arm uses the highest value at which it starts and serves the batch
($0.87$--$0.98$). The fraction reserves memory for the weights and the target KV pool;
CUDA-graph capture, activations and the draft worker's buffers must fit in the rest.
Native's draft worker must also allocate a full-length draft KV pool, so native reaches its
limit at a lower fraction; with the prefix cache on, it fails to start at $0.90$ while
allocating that pool, a single $6.25$\,GiB buffer (App.~\ref{app:online}). The windowed
pool holds only $n_{\text{sink}}{+}W{+}d$ slots per request, so the windowed arm can give
more memory to the target KV. The frontier therefore compares each arm's best feasible serving configuration, not equal fractions. On Qwen-122B ($0.90$) and Nemotron ($0.87$) every arm uses the same fraction.

\paragraph{Why the extra Qwen-35B request is a fair comparison.} Both arms run on
the same GPU with the same weights and start from the same $177.6$\,GiB of free memory; the
memory fraction only sets how SGLang splits it. The difference is the draft KV pool.
Native's draft pool stores every token of the target pool. At its highest working fraction,
$0.92$, the target pool holds $5.77$M tokens, the draft pool takes $11.0$\,GiB, and
$3.1$\,GiB is left. A sixth 1M request needs a ${\approx}6.24$M-token pool. Each extra token
costs native $22$\,KiB: $20$\,KiB of target KV ($10$ full-attention layers $\times$ $2$ KV
heads $\times$ $256$ dims $\times$ K and V in bf16) plus $2$\,KiB of draft KV (one layer). It would
need ${\approx}9.9$\,GiB more, over three times what is left. A higher fraction does not help:
at $0.98$ native fails to start, because its draft pool must fit in the $2\%$ outside the
fraction. The ring needs only $4{,}103$ draft slots ($8$\,MiB) per request; its pool
reserves them for SGLang's $39$ maximum running requests ($160{,}017$ slots, $0.3$\,GiB in all). The windowed
arm therefore runs at $0.98$ with a $6.28$M-token pool, seats six, and still has $1.9$\,GiB
free after allocating its draft pool.
The speedup does not depend on that extra memory: with windowing at $0.90$ ($5.60$M tokens),
below native's $0.92$, both arms seat five and queue the sixth, and windowing is still
$1.21\times$ faster per token at $B{=}5$ ($9.83$ vs.\ $11.89$\,ms). The sixth request is
therefore what the ring's freed draft memory buys on the same GPU.

Other settings: \texttt{page-size 1}, FlashInfer target attention, CUDA graphs on, and
MTP speculation (SGLang's \texttt{NEXTN} algorithm) with \texttt{num-steps}$=\gamma=d{-}1$ and \texttt{eagle-topk 1}. All
generation arms use repetition penalty $1.15$; without it, long greedy enumeration answers
can fall into repetition loops, which are easy to predict and inflate acceptance. Native
MTP uses the FlashInfer draft backend; Windowed-MTP uses the Triton draft backend, where
we implemented the ring remap, with $W{=}4032$ and $n_{\text{sink}}{=}64$.

\paragraph{Measurement protocol.} Every input is $1{,}040{,}000$ tokens. Each run does one
tiny warmup generation, flushes the cache, then a timed prefill and decode. TPOT comes from
SGLang's per-step log (\texttt{decode-log-interval 1}). Line $i$ reports the batch-mean
accept length $\AL_i$ and the aggregate throughput $X_i$ over that step, so the step's wall
time is $\Delta t_i=B_{\text{eff}}\AL_i/X_i$. Over steady steps (those at the full batch
$B_{\text{eff}}$ after a two-step ramp, clipped to $0.4$--$2.5\times$ the median
$\Delta t_i$) we set $\text{TPOT}=\operatorname{med}_i(\Delta t_i)/\overline{\AL}$, with
$\overline{\AL}$ taken over the same steps. At $B{=}1$ this matches the direct estimate
$B_{\text{eff}}\sum_i\Delta t_i/\sum_i N_i$ ($N_i=B_{\text{eff}}\AL_i$ committed tokens)
over all steady steps to within $0.6\%$ on every cell of the single-prompt sweep, and the
clip removes no steps. For $B{>}1$ sweeps (Figs.~\ref{fig:pareto} and~\ref{fig:tp}) we use that
direct estimate, unclipped, over the first full-batch stretch: from the first step at full
concurrency (after the ramp) to the first request finishing or being admitted. Later steps
mix KV teardown and re-admission prefill into the step log. The stretch keeps a one-time
${\sim}200$\,ms stall that speculative runs show early in decode; on short Nemotron decodes
it adds up to ${\sim}30\%$ to TPOT.

Fixed-batch runs use real-input QA, stopping at EOS or $512$ tokens. Batch and Pareto sweeps
(Figs.~\ref{fig:pareto} and~\ref{fig:tp}) copy one record across the batch with the radix
cache off, so each request holds its own full 1M KV. The online run is the exception
(App.~\ref{app:online}).

\paragraph{Statistical protocol.} Each cell of Table~\ref{tab:hero} runs $16$ prompts
per task (records $0$--$15$ of the seeded generator; Code-QA has $15$), one seeded run
each, and reports the median over prompts; the dense baseline runs the same prompts. Table~\ref{tab:alci} gives 95\% bootstrap
intervals ($10^4$ resamples) for the median $\AL$ ratio and the median nat/win. Nemotron
answers NIAH-s, VT, Code-QA and BABILong in $4$--$12$ tokens ($1$--$2$ verify steps), which
leaves no steady decode to time, so those cells are omitted. Within one run, TPOT is a
median over the steady steps, and a repeated greedy run reproduces it within about
$1\%$. Fig.~\ref{fig:alpha} pools the acceptance histograms of these same runs over all
prompts and inputs. Decode steps within one prompt are correlated, so its 95\% intervals come
from a cluster bootstrap over prompts, stratified by input. Unless stated otherwise, the other fixed-batch analyses use one prompt (record $0$, the first of the $16$).

\paragraph{The windowing change (pseudocode).} Windowed-MTP changes only a few lines in
the code that builds the draft's per-request KV index. The target verification path is
untouched.
{\footnotesize
\begin{verbatim}
# draft step, req r, seq len S:
if WINDOW and S > n_sink + W:
    keep = range(0, n_sink)   # sink
         + range(S - W, S)    # window
else:
    keep = range(0, S)        # short
kv_index = req_to_token[r, keep]
# RING: slot = ring_base(r) + logical_pos
#       modulus = n_sink + W + d
\end{verbatim}
}
RoPE is already applied to the cached keys, so shrinking the index does not change any
key's position. No new kernel or attention mask is needed, and the path is CUDA-graph safe.

\paragraph{The ring buffer.} With the ring on, the draft's KV pool holds
$n_{\text{sink}}{+}W{+}d$ slots per request instead of one slot per token. With
$s{=}n_{\text{sink}}$ and $M{=}W{+}d$, a token at position $p$ of request $r$ is stored at a
slot that depends only on $(r,p)$:
\[
  \text{slot}(r,p) = r\,(s{+}M) +
  \begin{cases} p, & p < s,\\
  s + (p-s) \bmod M, & p \ge s.\end{cases}
\]
The sink positions are written once; the rest wrap around a band of $W{+}d$ slots. A draft
step reads the sink plus the last $W$ positions and writes at most $d$ new ones, a span of
at most $W{+}d$ consecutive positions, so all map to distinct slots and the draft never
reads an overwritten entry. Rejection needs no rollback: beyond the committed sequence,
the draft reads only positions written in the current step, and a rejected token's slot is
overwritten when the committed token is written there. The target's KV, allocator and
position map are unchanged. The patch requires the Triton draft backend and refuses tree
drafts ($\texttt{eagle-topk}{>}1$) with page size above one.

\section{Latency-model fit}
\label{app:latency}

\paragraph{Acceptance-free per-iteration time.}
We fit time per iteration rather than TPOT, so acceptance does not enter. The scheduler
reports each iteration's accepted tokens $\AL_i$ and throughput $X_i$, so its wall time is
$T_{\text{iter},i}=\AL_i/X_i=\AL_i\cdot\text{TPOT}_i$, in which $\AL$ cancels. It depends
on the hardware and context length, not on the workload.

\paragraph{The fit.}
Following Eq.~\ref{eq:tpot}, $T_{\text{iter}}(\gamma)=c+\gamma\,\tfwd$ with
$c\equiv\tverify+\tctx$. We sweep $\gamma$ and fit this line by least squares over
$\gamma\in[4,7]$ ($d{=}5$--$8$; Qwen-122B over $\gamma\in[4,6]$, since $d{=}8$ does not fit
at 1M), after the one-time verify jump at $d{=}5$ explained below. On
\texttt{niah\_multiquery\_enum} (Table~\ref{tab:cost}) the fit is tight ($R^2{\ge}0.97$ for
both arms), and windowing lowers both terms.

The intercept falls by $\Delta c\approx5.1$--$6.1$\,ms. Windowing leaves $\tverify$
unchanged on the Qwen models, so this is mostly the draft's index build. Two kernels build
that index over the whole context,
\texttt{generate\_\allowbreak draft\_\allowbreak decode\_\allowbreak kv\_\allowbreak indices}
(${\approx}3.5$\,ms per iteration) and
\texttt{create\_\allowbreak flashinfer\_\allowbreak kv\_\allowbreak indices}
(${\approx}1$\,ms), and both drop to nearly zero under windowing
(Table~\ref{tab:kernelattr}). This saving depends on the implementation. The slope falls
by $0.3$--$0.5$\,ms per forward (${\approx}1.9$--$2.9$\,ms at $\gamma{=}6$); it comes from
each forward reading a bounded window, so it holds on any backend.

\paragraph{If native's index build cost nothing.} To bound how much a leaner index would
erase, we give native the whole intercept drop for free:
$(\text{step}^{\text{nat}}{-}|\Delta c|)/\text{step}^{\text{win}}$. This is native's best
case: a bandwidth-bound build of the ${\approx}4$\,MB index of a 1M request takes a few
microseconds, and $\Delta c$ also contains Nemotron's $1.2$\,ms bookkeeping drop, which is
not index work. Windowing still wins per step: $1.16\times$ on Qwen-35B ($21.3$ vs.\
$18.3$\,ms), $1.07\times$ on Qwen-122B ($28.4$ vs.\ $26.5$) and $1.07\times$ on Nemotron
($27.6$ vs.\ $25.8$; $1.12\times$ if the bookkeeping drop is kept). What remains is the
per-forward read saving.

\paragraph{The verify jump at $d{=}5$.} In a finer sweep over $\gamma$ (to $\gamma{=}6$ on Qwen-122B and $7$ on Nemotron, the deepest that fit a 1M request), native
$T_{\text{iter}}$ on Qwen-122B and Nemotron jumps between $\gamma{=}3$ and $4$
($24.8{\to}32.2$ and $24.5{\to}30.7$\,ms). An nsys profile places ${\approx}90\%$ of it in
the verify attention: going from $4$ to $5$ query rows pushes FlashInfer's paged-prefill
kernel past a tile boundary, roughly doubling its cost per call ($593{\to}1133\,\mu$s).
This adds ${\approx}6$\,ms once to $\tverify$, and so to $c$. It leaves the slope unchanged
and affects both arms equally, so it cancels in $\Delta c$. Fitting only after the jump
keeps every point, including $d{=}7$, on the same side of it.

Table~\ref{tab:kernelattr} lists the largest kernels in $\tdraft$ from the nsys trace behind Table~\ref{tab:iterphase}.
These are GPU-busy times, which exclude idle and CPU time and so sum to less than $\tdraft$.

\begin{table}[!htb]
  \centering
  \caption{Largest components of $\tdraft$ (nsys GPU-busy, $d{=}7$, 1M, $B{=}1$,
  $\mu$s/iter). The full-context index kernels collapse under windowing. Qwen-122B's slower
  windowed draft-extend is Triton's ragged extend kernel, not the window
  (App.~\ref{app:scope}).}
  \label{tab:kernelattr}
  \footnotesize
  \adjustbox{max width=\linewidth}{%
  \begin{tabular}{lrrr}
    \toprule
    draft kernel ($\mu$s/iter) & q35 & q122 & nem \\
    \midrule
    \multicolumn{4}{l}{\emph{native (FlashInfer draft, full pool)}}\\
    \quad \texttt{generate\_draft\_decode\_kv\_indices} & 3588 & 3655 & 2891 \\
    \quad \texttt{create\_flashinfer\_kv\_indices} (draft) & 1047 & 1068 & 1032 \\
    \quad draft-decode attn ($\times6$) & 1608 & 1598 & 853 \\
    \quad draft-extend attn & 423 & 818 & 404 \\
    \midrule
    \multicolumn{4}{l}{\emph{windowed+ring (Triton draft, compact pool)}}\\
    \quad \texttt{generate\_draft\_decode\_kv\_indices} & 3.0 & 3.1 & 3.1 \\
    \quad ring index kernel & 2.0 & 2.2 & 2.0 \\
    \quad draft-decode attn ($\times6$) & 348 & 976 & 149 \\
    \quad draft-extend attn & 550 & 2908 & 143 \\
    \bottomrule
  \end{tabular}}
\end{table}

\section{Predicting when native speculation loses}
\label{app:predict}

\S\ref{sec:results} shows that native speculation can be slower than none. The
break-even $\AL$ of Eq.~\ref{eq:breakeven} uses only acceptance-free step times, so a
server can compute it before any request arrives. Here we calibrate it on one input and
check its calls on the rest.

\paragraph{The test.} For each input we divide native's $d{=}7$ step time, measured on
\texttt{niah\_multiquery\_enum} only (Table~\ref{tab:cost}), by that input's dense TPOT to
get its break-even $\AL$, and flag native as losing when its measured $\AL$ falls below
it (Table~\ref{tab:predict}). $17$ of the $20$ cells are held out.

\begin{table}[!ht]
  \centering
  \caption{Break-even check at $d{=}7$, 1M, $B{=}1$. Break-even $\AL$ is native's step
  time on \texttt{niah\_multiquery\_enum} ($\dagger$; Table~\ref{tab:cost}) over each
  input's dense TPOT; $\AL$ and dense are from Table~\ref{tab:hero}. \emph{inverts}: native
  slower than dense, predicted (p: $\AL$ below break-even) vs.\ measured (m). \textbf{Bold}: the three false positives; there are no misses.}
  \label{tab:predict}
  \small
  \setlength{\tabcolsep}{3.5pt}
  \begin{tabular}{llccc}
    \toprule
    model & input & $\AL$ & break-even & inverts (p/m) \\
    \midrule
    \multirow{8}{*}{q35}
      & NIAH-s & 5.07 & 2.68 & -- / -- \\
      & NIAH-mv & 5.42 & 2.68 & -- / -- \\
      & NIAH-mq$^\dagger$ & 4.45 & 2.70 & -- / -- \\
      & VT & 4.18 & 2.72 & -- / -- \\
      & CWE & 2.74 & 2.87 & \textbf{inv} / -- \\
      & FWE & 2.85 & 2.86 & \textbf{inv} / -- \\
      & Code-QA & 3.30 & 2.65 & -- / -- \\
      & BABILong & 3.07 & 2.74 & -- / -- \\
    \midrule
    \multirow{8}{*}{q122}
      & NIAH-s & 4.04 & 3.62 & -- / -- \\
      & NIAH-mv & 4.81 & 3.62 & -- / -- \\
      & NIAH-mq$^\dagger$ & 4.76 & 3.61 & -- / -- \\
      & VT & 3.82 & 3.59 & -- / -- \\
      & CWE & 3.28 & 3.89 & inv / inv \\
      & FWE & 3.45 & 3.93 & inv / inv \\
      & Code-QA & 3.16 & 3.60 & inv / inv \\
      & BABILong & 3.64 & 3.70 & \textbf{inv} / -- \\
    \midrule
    \multirow{4}{*}{nem}
      & NIAH-mv & 2.86 & 2.92 & inv / inv \\
      & NIAH-mq$^\dagger$ & 3.63 & 2.91 & -- / -- \\
      & CWE & 1.64 & 3.04 & inv / inv \\
      & FWE & 1.64 & 3.03 & inv / inv \\
    \bottomrule
  \end{tabular}
\end{table}

\paragraph{Accuracy.} All $6$ cells where native is slower than dense are flagged, and the
$3$ false positives are near-ties: q35 CWE and FWE (native $1.05\times$ and $1.07\times$
dense) and q122 BABILong ($1.00\times$, a tie). This is the less costly error: a server would turn speculation down slightly
early rather than keep a losing configuration. The question only arises for native
speculation: windowing's per-step saving does not depend on acceptance, and Windowed-MTP
is ahead of native on every cell.

\paragraph{Where the error comes from.} The calls assume step time is the same on every
input. It nearly is: on $13$ of the $20$ cells native's median step is within $3\%$ of
the calibration value. The exceptions are the aggregation inputs (CWE and FWE), whose step
is $7$--$9\%$ shorter, likely because the draft-extend step scales with accepted tokens,
and q35 BABILong ($3.1\%$). A shorter step lowers the true break-even, so these inputs are
flagged slightly early; the two q35 false positives come from this, and no call is missed.

\paragraph{Other depths.} With native's step predicted by the line of Table~\ref{tab:cost}
instead of measured, the test on the other inputs' single-prompt depth sweeps flags all $20$
of $53$ cells at $d{=}5$--$8$ where native loses ($4$ near-tie false positives,
$1.00$--$1.06\times$). The line does not carry past a tile-boundary jump ($d{=}2{\to}3$ on
all models, $d{=}9$ on Qwen-35B), where a separate fit is needed.

\section{Per-position and decision-invariance details}
\label{app:mech}

Table~\ref{tab:alpha} gives the numbers behind Fig.~\ref{fig:alpha}, pooled over all
prompts and inputs. On the Qwen models native and windowed differ by at most $0.04$ at any
position (Qwen-122B at $j{=}2$, $0.78{\to}0.74$; Qwen-35B at $j{=}3$, $0.77{\to}0.73$), and
$\alpha_1$ by at most $0.02$. Nemotron's windowed draft accepts more at every position. Deep
positions have the fewest samples, most visibly on Nemotron, where about $200$ steps per draft
reach $j{=}6$. Single inputs vary more: the largest per-input change on Qwen is $-0.18$
(Qwen-35B NIAH-mv, $j{=}2$).

\begin{table}[!htb]
  \centering
  \caption{Conditional acceptance $\alpha_j$ by draft position ($d{=}7$, 1M), pooled
  over all prompts and inputs of Table~\ref{tab:hero}; the numbers behind
  Fig.~\ref{fig:alpha}.}
  \label{tab:alpha}
  \small
  \adjustbox{max width=\linewidth}{\begin{tabular}{llcccccc}
    \toprule
    model & draft & $\alpha_1$ & $\alpha_2$ & $\alpha_3$ & $\alpha_4$ & $\alpha_5$ & $\alpha_6$ \\
    \midrule
        \multirow{2}{*}{q35}  & native & 0.77 & 0.75 & 0.77 & 0.77 & 0.77 & 0.79 \\
                          & window & 0.76 & 0.73 & 0.73 & 0.74 & 0.80 & 0.79 \\
    \multirow{2}{*}{q122} & native & 0.80 & 0.78 & 0.75 & 0.78 & 0.81 & 0.81 \\
                          & window & 0.78 & 0.74 & 0.75 & 0.78 & 0.81 & 0.80 \\
    \multirow{2}{*}{nem}  & native & 0.45 & 0.52 & 0.63 & 0.38 & 0.38 & 0.20 \\
                          & window & 0.52 & 0.56 & 0.66 & 0.43 & 0.42 & 0.40 \\
    \bottomrule
  \end{tabular}}
\end{table}

\paragraph{Direct decision-invariance probe.}
Acceptance depends on whether windowing changes the draft's \emph{choice} of token. In the
native run ($1$M, $d{=}7$, \texttt{niah\_multiquery\_enum}), at each decode step we rerun
the draft head with the same hidden state, position and seed but with its KV cut to the
sink and window, and compare the two top-1 proposals. Over $275$--$512$ draft-head forwards (each speculative step runs several; the probe fires on every draft decode forward past $4$K tokens of context and stops at $512$ or when the standard $512$-token generation ends, Table~\ref{tab:top1}) the top-1 is unchanged $86$--$94\%$ of the time. For Qwen-35B and
Nemotron the flips fall at near-ties, where the full model's top-two logit gap is
${\approx}0.5$, against $3$--$5$ on average. The size of the internal change does not
decide this: under windowing Nemotron picks about half of its MoE experts differently
(router-overlap Jaccard ${\approx}0.5$), yet keeps its top-1 $86\%$ of the time.

Qwen-122B is the exception. Its flips fall at confident positions (gap $6.8$ against an
average of $7.2$): this strongest drafter uses distant context to tell needles apart, which
explains its acceptance loss on this input ($5.57{\to}4.74$ on the probed prompt; median
ratio $0.89$ over $16$ prompts, Table~\ref{tab:alci}). The
target verifies every token, so this costs latency, never correctness. The probe is per
step; whole-generation effects are measured by the windowed runs in Table~\ref{tab:hero}.

\begin{table}[!htb]
  \centering
  \caption{Decision-invariance probe (native draft, in-run A/B, eager; 1M, $B{=}1$,
  $d{=}7$, \texttt{niah\_multiquery\_enum}). \emph{probes}: draft-head decode forwards compared, within the run's $512$-token generation. \emph{agree}: fraction of probes whose windowed
  top-1 equals the full-context top-1. \emph{margin}: the full model's top-1/top-2 logit
  gap, over all steps and over flipped steps only.}
  \label{tab:top1}
  \small
  \adjustbox{max width=\linewidth}{\begin{tabular}{lccccc}
    \toprule
    model & probes & agree & flip & margin (all) & margin (flips) \\
    \midrule
    q35  & 512 & 0.94 & 0.06 & 5.16 & 0.49 \\
    q122 & 465 & 0.87 & 0.13 & 7.18 & 6.80 \\
    nem  & 275 & 0.86 & 0.14 & 3.02 & 0.54 \\
    \bottomrule
  \end{tabular}}
\end{table}

\paragraph{Why the decision holds.}
Both draft families sit on recurrent-hybrid models (Gated-DeltaNet for Qwen, Mamba2 for
Nemotron, each with a few full-context GQA layers). Their recurrent layers already carry
long-range information, so the draft's distant reread mostly repeats what it has, consistent
with concurrent attention-drift findings (\S\ref{sec:background}). How far the
attention reaches depends on positional encoding (Table~\ref{tab:attnmass}). The Qwen heads
(length-extended RoPE) put ${\approx}75$--$80\%$ of their softmax mass inside the window,
and windowing changes their attention output by ${\sim}15\%$. The Nemotron head (NoPE) puts
only ${\approx}16\%$ inside and its output changes by over $100\%$, yet its top-1 is
equally preserved: the Mamba2 state already encodes that context, and the attention
output's L2 norm is about half that of the hidden state it is added to
($\lVert W_o o\rVert/\lVert h_t\rVert{\approx}0.5$). The size of the change therefore does
not predict the token choice (Table~\ref{tab:top1}).

\begin{table}[!htb]
  \centering
  \caption{Attention mass of the native draft at the operating window ($W{=}4032$, sink
  $64$; 1M, $B{=}1$, $d{=}5$, \texttt{niah\_multiquery\_enum}). in-win mass: softmax mass
  inside the window; $\wperturb$: relative $L_2$ change of the read-out $o{=}\sum_j p_j v_j$
  when the far tail is dropped; tail out-frac: share of $\lVert o\rVert$ from far keys.}
  \label{tab:attnmass}
  \small
  \begin{tabular}{lccc}
    \toprule
    model & in-win mass & $\wperturb$ & tail out-frac \\
    \midrule
    q35  & 0.748 & 0.139 & 0.246 \\
    q122 & 0.805 & 0.169 & 0.184 \\
    nem  & 0.160 & 1.059 & 0.822 \\
    \bottomrule
  \end{tabular}
\end{table}

Windowing could lose acceptance where distant attention is \emph{concentrated} on a few
tokens that matter. The largest loss on this prompt is $15\%$ (Qwen-122B), and across all
tasks the largest task-level median loss is $19\%$ (Qwen-35B NIAH-mv, not probed this way); the cost saving outweighs it on every
cell. This is interpretation, not proof. Swapping the distant content in a controlled way
would separate two explanations, distant attention diluting the prediction and
length-extended RoPE degrading at long range; we leave that to future work.

\section{Ablations and scope notes}
\label{app:scope}

\paragraph{The backend is not the confound.}
Native MTP runs on FlashInfer, but we implemented the ring on the Triton draft backend, so the gain
could in principle come from the backend. It does not: Triton is a handicap. At a matched
point (Table~\ref{tab:backend}), moving native to Triton makes its full-context
draft-extend much slower, yet Windowed-MTP on Triton still beats native on FlashInfer. One
caveat: Triton's ragged draft-extend grows with accept length, unlike FlashInfer's padded
\texttt{BatchPrefill}, which can eat into the gain at high acceptance
(Table~\ref{tab:kernelattr}). We have not built a FlashInfer ring, so its performance remains unmeasured.

\begin{table}[!htb]
  \centering
  \caption{Backend ablation at a matched point (q35, $d{=}5$, 1M, CWE, $B{=}1$). Native on
  Triton is a deliberate stress test, not an operating point: Triton makes native's TPOT ${\sim}6\times$ worse, yet windowed on Triton beats native on
  FlashInfer.}
  \label{tab:backend}
  \small
  \begin{tabular}{llc}
    \toprule
    draft path & backend & TPOT (ms) \\
    \midrule
    native            & FlashInfer & 8.13 \\
    native            & Triton     & 50.59 \\
    Windowed-MTP (ring buffer) & Triton   & \textbf{5.37} \\
    \bottomrule
  \end{tabular}
\end{table}

\paragraph{Robustness to KV-cache quantization.}
Windowing cuts the \emph{number} of KV entries the draft reads (about $99\%$ fewer at 1M);
the KV dtype sets the bytes per entry. Under fp8 the draft still skips ${\approx}99\%$ of
its KV bytes, but the absolute time saved scales with bytes per entry, so lower-precision
KV shrinks the relative gain to the extent decode is KV-bandwidth bound. A sweep on
\texttt{niah\_multiquery\_enum} confirms this (per-step ratio
$\text{step}^{\mathrm{nat}}/\text{step}^{\mathrm{win}}$, in which acceptance cancels):
\begin{center}
\footnotesize
\setlength{\tabcolsep}{5pt}
\begin{tabular}{lcccc}
\toprule
 & \multicolumn{2}{c}{261K} & \multicolumn{2}{c}{1M} \\
\cmidrule(lr){2-3}\cmidrule(lr){4-5}
model & bf16 & fp8 & bf16 & fp8 \\
\midrule
Qwen-35B  & $1.28\times$ & $1.20\times$ & $1.46\times$ & $1.22\times$ \\
Qwen-122B & $1.23\times$ & $1.14\times$ & $1.29\times$ & $1.15\times$ \\
Nemotron  & $1.18\times$ & $1.16\times$ & $1.26\times$ & $1.26\times$ \\
\bottomrule
\end{tabular}
\end{center}
Windowing wins every cell ($1.14$--$1.26\times$ under fp8). This is a separate run, so its bf16 column differs slightly from Table~\ref{tab:cost}. fp8 shrinks the Qwen gains (at 1M,
$1.46{\to}1.22\times$ and $1.29{\to}1.15\times$), while Nemotron, which reads less KV per
token, is unchanged ($1.26\times$). Verification still preserves the output distribution,
and the reclaimable fraction of the draft pool is unchanged; only its size in GB scales.
Our main numbers fix KV storage to bf16 for a controlled comparison and evaluate fp8 KV
separately; our stack's fp8 KV path also lacks a fused attention kernel, which inflates
absolute fp8 latency.

\paragraph{Compatibility with tree speculation.}
We evaluate chain speculation (\texttt{eagle-topk}${=}1$). Windowing changes only which KV
the draft reads, not the candidate tree, so it applies to tree speculation unchanged; only
the ring grows, to $n_{\text{sink}}{+}W{+}d{\cdot}k$ slots for width $k$, still far below a
full-length draft pool. Measuring it is future work.

\paragraph{Pruning the draft's prefill to lower TTFT.}
The draft builds its own KV over the prompt, one pass per prefill chunk, but under
windowing it only reads the sink and the last $W$ tokens. Skipping the intermediate chunks
would shrink draft prefill from $O(S)$ to $O(W{+}n_{\text{sink}})$, removing almost all of
it at 1M and lowering TTFT without touching the target or the output distribution. The
kernel work is future work.

\section{Online serving under continuous arrivals}
\label{app:online}

\paragraph{Workload.} Qwen-35B at TP2 under SGLang's continuous-batching scheduler with the
prefix cache on. Requests arrive as a Poisson process at rate $\lambda$ (req/s), with input
lengths of 131K, 262K, 524K and 1.04M tokens in a $4{:}3{:}2{:}1$ ratio, drawn from the RULER
QA corpora (\texttt{niah\_multiquery\_enum}, \texttt{cwe}, \texttt{fwe\_enum}). Half of the
requests ask a new question about an already-served context (measured prefix hit rate
$0.69$). We discard $8$ warmup requests and report the next $32$.

\paragraph{Controls.} (1)~One seed fixes the whole trace, so request $i$ is the same work in
every arm and requests can be compared in pairs. (2)~With \texttt{ignore\_eos} every request
generates exactly $4096$ tokens: at the main runs' $512$-token cap decode is only
$20$--$27\%$ of request time next to a 1M prefill, against $45$--$74\%$ at $4096$. (3)~All arms use \texttt{mem-fraction-static} $0.85$;
at $0.90$ native fails to start while allocating its full-length draft pool (a single
$6.25$\,GiB buffer), while dense and windowed run.

\paragraph{Results.} Windowing's paired median speedup is $1.15$--$1.21\times$ over native
and $2.0$--$2.2\times$ over dense (Table~\ref{tab:onlinepaired}); the mean gain is larger
($1.2$--$1.8\times$), so it is concentrated in long requests. Under a $25$\,ms p95-TPOT SLO
on the longest bucket, windowing sustains $0.05$~req/s and both baselines only $0.02$
(Table~\ref{tab:onlineslo}); windowing's worst bucket ($22.8$\,ms) is below native's best
($36.0$\,ms). With two rates and four 1.04M requests, this is a separation, not a capacity estimate. Online $\AL$ is
$3.61$--$3.77$ for native and $3.15$--$3.42$ for windowed: windowing wins on cheaper steps,
not on acceptance.

\begin{table}[!htb]
  \centering
  \caption{Paired per-request decode latency, online trace (Qwen-35B TP2, $32$
  steady-state requests). wins: requests on which windowing is strictly faster.}
  \label{tab:onlinepaired}
  \small
  \setlength{\tabcolsep}{5pt}
  \adjustbox{max width=\linewidth}{\begin{tabular}{lccc|cc}
    \toprule
    & \multicolumn{3}{c|}{mean ms/token} & \multicolumn{2}{c}{paired median (wins)} \\
    $\lambda$ & dense & native & win & vs.\ dense & vs.\ native \\
    \midrule
    $0.02$ & 9.92  & 6.50  & \textbf{5.34}  & $2.02\times$ (29/32) & $1.15\times$ (26/32) \\
    $0.05$ & 26.26 & 18.87 & \textbf{10.51} & $2.23\times$ (32/32) & $1.21\times$ (28/32) \\
    \bottomrule
  \end{tabular}}
\end{table}

\begin{table}[!htb]
  \centering
  \caption{Per-bucket p95 TPOT (ms) at $\lambda{=}0.05$ and the highest tested rate
  ($0.02$ or $0.05$) sustained under each SLO. Buckets are small ($n{=}4$ at 1.04M).}
  \label{tab:onlineslo}
  \small
  \setlength{\tabcolsep}{5pt}
  \adjustbox{max width=\linewidth}{\begin{tabular}{lcccc|cc}
    \toprule
    & \multicolumn{4}{c|}{p95 TPOT (ms) by input length} &
      \multicolumn{2}{c}{max sustained $\lambda$} \\
    arm & 131K & 262K & 524K & 1.04M & 25\,ms & 50\,ms \\
    \midrule
    dense    & 39.3 & 41.0 & 43.6 & 39.0 & $0.02$ & $0.05$ \\
    native   & 38.4 & 38.8 & 41.3 & 36.0 & $0.02$ & $0.05$ \\
    windowed & \textbf{18.2} & \textbf{21.2} & \textbf{12.1} & \textbf{22.8}
             & $\mathbf{0.05}$ & $0.05$ \\
    \bottomrule
  \end{tabular}}
\end{table}

\paragraph{The ceiling is set by prefill.} From $0.075$ to $0.15$~req/s offered, no arm keeps up:
served rates level off at $0.054$--$0.063$~req/s and p95 TTFT grows from $155$ to $326$\,s.
Windowing raises the rate a server can handle under a decode SLO, not peak throughput,
which needs prefill work such as the draft-prefill pruning of App.~\ref{app:scope}.

\section{Variation across prompts}
\label{app:tables}

Table~\ref{tab:alci} gives the per-prompt spread for every cell of Table~\ref{tab:hero}.
Native is faster on $18$ of $318$ prompts, and no cell's median favors native. The per-step
speedup is nearly constant across prompts (cell medians within $3\%$ of
Table~\ref{tab:cost}), so the spread in end-to-end speedup comes from acceptance.

\begin{table}[!htb]
  \centering
  \caption{Per-prompt spread ($d{=}7$, 1M, $B{=}1$): medians of
  $\AL^{\mathrm{win}}/\AL^{\mathrm{nat}}$ and nat/win TPOT, 95\% bootstrap CIs ($10^4$
  resamples); range: min--max nat/win; nat faster: prompts with nat/win $<1$.}
  \label{tab:alci}
  \small
  \setlength{\tabcolsep}{4pt}
  \adjustbox{max width=\linewidth}{\begin{tabular}{llccccc}
    \toprule
    model & input & $n$ & $\AL$ ratio [95\% CI] & nat/win [95\% CI] & range & nat faster \\
    \midrule
    \multirow{8}{*}{q35}
      & NIAH-s & 16 & $0.94\,[0.90,0.98]$ & $1.35\,[1.30,1.39]$ & $1.21$--$1.46$ & 0 \\
      & NIAH-mv & 16 & $0.81\,[0.75,0.86]$ & $1.15\,[1.05,1.17]$ & $0.97$--$1.29$ & 1 \\
      & NIAH-mq & 16 & $0.96\,[0.92,1.02]$ & $1.34\,[1.29,1.43]$ & $1.19$--$1.63$ & 0 \\
      & VT & 16 & $0.84\,[0.81,0.98]$ & $1.21\,[1.16,1.41]$ & $1.05$--$1.71$ & 0 \\
      & CWE & 16 & $1.04\,[0.96,1.10]$ & $1.45\,[1.35,1.53]$ & $0.60$--$2.07$ & 1 \\
      & FWE & 16 & $1.11\,[1.01,1.28]$ & $1.56\,[1.45,1.83]$ & $1.26$--$2.34$ & 0 \\
      & Code-QA & 15 & $0.95\,[0.90,1.03]$ & $1.37\,[1.30,1.50]$ & $1.12$--$1.73$ & 0 \\
      & BABILong & 16 & $0.90\,[0.85,1.03]$ & $1.28\,[1.19,1.43]$ & $0.69$--$2.25$ & 3 \\
    \midrule
    \multirow{8}{*}{q122}
      & NIAH-s & 16 & $0.93\,[0.91,0.99]$ & $1.21\,[1.18,1.28]$ & $0.96$--$1.43$ & 1 \\
      & NIAH-mv & 16 & $0.87\,[0.82,0.92]$ & $1.14\,[1.06,1.22]$ & $1.00$--$1.45$ & 2 \\
      & NIAH-mq & 16 & $0.89\,[0.86,0.94]$ & $1.16\,[1.11,1.23]$ & $0.96$--$1.66$ & 1 \\
      & VT & 16 & $0.96\,[0.89,1.08]$ & $1.25\,[1.16,1.40]$ & $1.08$--$1.81$ & 0 \\
      & CWE & 16 & $0.98\,[0.93,1.09]$ & $1.27\,[1.20,1.42]$ & $1.14$--$1.65$ & 0 \\
      & FWE & 16 & $1.04\,[0.98,1.09]$ & $1.34\,[1.25,1.39]$ & $1.11$--$1.86$ & 0 \\
      & Code-QA & 15 & $1.01\,[0.93,1.05]$ & $1.31\,[1.23,1.41]$ & $0.97$--$1.67$ & 1 \\
      & BABILong & 16 & $0.90\,[0.76,0.96]$ & $1.14\,[0.98,1.25]$ & $0.70$--$1.73$ & 4 \\
    \midrule
    \multirow{4}{*}{nem}
      & NIAH-mv & 16 & $0.97\,[0.93,1.00]$ & $1.26\,[1.19,1.30]$ & $1.08$--$1.91$ & 0 \\
      & NIAH-mq & 16 & $1.00\,[0.94,1.07]$ & $1.27\,[1.20,1.36]$ & $0.66$--$3.19$ & 1 \\
      & CWE & 16 & $1.15\,[1.04,1.51]$ & $1.47\,[1.33,1.97]$ & $1.09$--$2.25$ & 0 \\
      & FWE & 16 & $1.19\,[0.92,1.49]$ & $1.53\,[1.19,1.97]$ & $0.71$--$3.31$ & 3 \\
    \bottomrule
  \end{tabular}}
\end{table}

\end{document}